%% file: main.tex
\documentclass{Definitions/ieeeaccess}
\usepackage{cite}
\usepackage{amsmath,amssymb,amsfonts}
\usepackage{algorithmic}
\usepackage{graphicx}
\usepackage{textcomp}
\usepackage[ruled,vlined]{algorithm2e}
\usepackage{algorithmic}
\usepackage{amsmath}
\usepackage{caption,setspace}
\usepackage{subcaption}
\usepackage{booktabs}
\usepackage{scalerel}
\usepackage{tikz}
\usetikzlibrary{svg.path}
\usepackage{silence}
\def\BibTeX{{\rm B\kern-.05em{\sc i\kern-.025em b}\kern-.08em
    T\kern-.1667em\lower.7ex\hbox{E}\kern-.125emX}}
\DeclareMathOperator*{\argmax}{argmax} 
\DeclareMathOperator*{\argsort}{argsort} 
\NewSpotColorSpace{PANTONE}
\AddSpotColor{PANTONE} {PANTONE3015C} {PANTONE\SpotSpace 3015\SpotSpace C} {1 0.3 0 0.2}
\SetPageColorSpace{PANTONE}%

\definecolor{orcidlogocol}{HTML}{A6CE39}
\tikzset{
  orcidlogo/.pic={
    \fill[orcidlogocol] svg{M256,128c0,70.7-57.3,128-128,128C57.3,256,0,198.7,0,128C0,57.3,57.3,0,128,0C198.7,0,256,57.3,256,128z};
    \fill[white] svg{M86.3,186.2H70.9V79.1h15.4v48.4V186.2z}                 svg{M108.9,79.1h41.6c39.6,0,57,28.3,57,53.6c0,27.5-21.5,53.6-56.8,53.6h-41.8V79.1z M124.3,172.4h24.5c34.9,0,42.9-26.5,42.9-39.7c0-21.5-13.7-39.7-43.7-39.7h-23.7V172.4z}                 svg{M88.7,56.8c0,5.5-4.5,10.1-10.1,10.1c-5.6,0-10.1-4.6-10.1-10.1c0-5.6,4.5-10.1,10.1-10.1C84.2,46.7,88.7,51.3,88.7,56.8z};
  }
}

\newcommand\orcidicon[1]{\href{https://orcid.org/#1}{\mbox{\scalerel*{
\begin{tikzpicture}[yscale=-1,transform shape]
\pic{orcidlogo};
\end{tikzpicture}
}{|}}}}
\usepackage{hyperref} 

\begin{document}
\bstctlcite{IEEEexample:BSTcontrol}

\title{An Autonomous Spectrum Management Scheme for Unmanned Aerial Vehicle Networks in Disaster Relief Operations}
\author{\uppercase{Alireza Shamsoshoara}\authorrefmark{1},
\uppercase{Fatemeh Afghah}\authorrefmark{1}, \uppercase{Abolfazl Razi}\authorrefmark{1}, \uppercase{Sajad Mousavi}\authorrefmark{1}, \uppercase{Jonathan Ashdown}\authorrefmark{2}, \uppercase{and Kurt Turk}\authorrefmark{2}}
\address[1]{School of Informatics, Computing, and Cyber Systems, Northern Arizona University, Flagstaff, AZ 86011 USA (e-mail: \{Alireza\_Shamsoshoara, Fatemeh.Afghah, Abolfazl.Razi, SajadMousavi\}@nau.edu)}
\address[2]{Air Force Research Laboratory, Rome, 
NY 13441 USA (e-mail: \{Jonathan.Ashdown, Kurt.Turk\}@us.af.mil)}
\tfootnote{Distribution A: Approved for Public Release, distribution unlimited.  Case Number 88ABW-2019-5665 on 20 Nov. 2019. The work of F. Afghah, J. Ashdown and K. Turk was supported by AFRL.}


\corresp{Corresponding author: Alireza Shamsoshoara (e-mail: alireza\_shamsoshoara@nau.edu).}
\begin{abstract}
This paper studies the problem of spectrum shortage in an unmanned aerial vehicle (UAV) network during critical missions such as wildfire monitoring, search and rescue, and disaster monitoring. Such applications involve a high demand for high-throughput data transmissions such as real-time video-, image-, and voice- streaming where the assigned spectrum to the UAV network may not be adequate to provide the desired Quality of Service (QoS). In these scenarios, the aerial network can borrow additional spectrum from the available terrestrial networks in trade of a relaying service for them.
We propose a spectrum sharing model in which the UAVs are grouped into two classes of \textit{relaying UAVs} that service the spectrum owner and the sensing UAVs that perform the disaster relief mission using the obtained spectrum. The operation of the UAV network is managed by a hierarchical mechanism in which a central controller assign the tasks of the UAVs based on their resources and determine their operation region based on the level of priority of impacted areas and then the UAVs autonomously fine-tune their position using a model-free reinforcement learning algorithm to maximize the individual throughput and prolong their lifetime. We analyze the performance and the convergence for the proposed method analytically and with extensive simulations in different scenarios.
\end{abstract}
\titlepgskip=-15pt

\maketitle
\input{Texfiles/intro.tex}
\input{Texfiles/related.tex}

\input{Texfiles/system.tex}
\input{Texfiles/algorithm.tex}
\input{Texfiles/simulation.tex}
\input{Texfiles/conclusion.tex}

\section*{Acknowledgment}
The authors acknowledge the U.S. Government's support in the publication of this paper. This material is in part based upon work funded by AFRL. Any opinions, findings and conclusions or recommendations expressed in this material are those of the author(s) and do not necessarily reflect the views of the US government or AFRL.

\bibliographystyle{IEEEtran}
\bibliography{main.bib}

\end{document}

%% file: Texfiles/intro.tex
\section{Introduction}\label{Introduction}


\PARstart{S}{everal} unique features of unmanned aerial vehicles (UAVs) including low cost and fast deployment, wide field of view, 3-dimensional movements, and aerial and terrestrial mapping make them very attractive for various applications such as disaster relief, military missions, wildfire monitoring, precision agriculture, and surveillance \cite{Gartner_UAV,mousavi2019use,sklivanitis2018airborne,piao2019automating, shamsoshoara2019overview,shamsoshoara2019ring,valehi2017maximizing,valehi2018online}.
Moreover, Internet of Things(IoTs) can benefit from UAVs as a relay to forward information to the legitimiate destination \cite{mozaffari2016unmanned,shamsoshoara2019survey}.
In certain applications namely military missions and disaster relief operations, there is a demand for high bandwidth communications to transmit sensed information to a data fusion center. Usually, the sensed information can be in the form of real-time video or high quality aerial images. The required data transmission rate depends on the dynamicity level of the operation field. For instance, the mission may involve short-term periods of time in which a very large bandwidth is required for real-time streaming that was not foreseen in the original spectrum allocation planning. In such cases, the pre-allocated spectrum to the UAV network may not be adequate to meet this demand. This need calls for new solutions to provide an additional spectrum for the  fleet of UAVs during such critical missions. 

In this study, we propose a solution to provide the  required additional spectrum for the UAV network by considering the network throughput and lifetime.
In common UAVs' operation fields in rural or urban settings, there usually exists a terrestrial network which owns the licensed spectrum also known as the primary network (PN).
This primary network can occasionally experience a low quality of communication due to shadowing, fading, or even the direct communication may be compromised because of the damages to infrastructure caused by natural disasters such as wildfires and earthquakes. Hence, utilizing a relay node can be beneficial for the primary or the terrestrial network to forward its messages to its legitimate receiver or improve its transmission quality by taking advantage of cooperative communication. For instance, a low-power IoT node in a remote area can benefit from a relay UAV when the direct communication to a distant destination requires a high transmission power. Our proposed spectrum sharing model exploits this opportunity to provide the additional required spectrum for the UAV networks during critical missions. In this model, one UAV serves as an aerial relay for the primary network in exchange for the required spectrum access that is borrowed for other UAVs in the network to complete their mission. Cooperative spectrum leasing models, also known as property-right models, in which the primary user as the owner of the spectrum leases a portion of its spectrum to the secondary users in exchange for some profit have been studied in several works \cite{Stanojev08,Afghah_CDC2013,Namvar15}. In these models, the primary network selects the appropriate relay nodes and determines an optimal time allocation strategy for spectrum access to maximize its own benefits. However, such unilateral spectrum leasing models are not appropriate to provide the extra demanded spectrum for the UAV networks during the critical missions as the spectrum sharing strategies are solely determined by the PU.

This paper investigates a disaster relief scenario where a fleet of UAVs collect imagery data (e.g., real-time video) and transmit the data to an emergency center. We develop a hierarchical cooperative spectrum sharing model for the UAVs to secure additional spectrum access from other available network. In this model, the task of the UAVs and the most appropriate region of operation for them is assigned by a central controller. This approach divides the grid surface into multiple regions and the controller unit determines the operation region of each UAV to optimize the throughput and  maximize the network lifetime considering the residual energy of each agent.  The controller takes into account the level of system's dynamicity in terms of several factors such as the variations in the environment conditions due to the disaster or the primary's location  to re-initiate the region assignment for UAVs if needed. Next, the UAVs perform a learning technique to find their optimal positions within their operation cell in a distributed manner. The UAVs are assumed to be autonomous in the sense that they can find their optimal path and location in a self-directed manner. They consider different factors such as the residual energy, position of the transmitter and the receiver and the throughput rate. A proof of convergence is provided along with simulation results to show the system performance in terms of throughput and lifetime maximization with minimizing the number of steps.

The contribution of this work is to develop a practical model for spectrum sharing in high-priority critical missions that can be locally managed by the users without the need for developing new regulations. The proposed hybrid approach  utilizes both centralized and distributed techniques to find the best solution for spectrum sharing and location optimization in a reasonable amount of time. The solution guarantees the optimum throughput and lifetime for both primary and secondary (i.e., UAV) networks. We like to note that the centralized decision making is not performed constantly but as needed depending on the level of variations in the system status. Hence, the center is not considered as a bottleneck for the system performance. Moreover, this hybrid solution can enable the spectrum sharing in scenarios where the central controller (i.e., the emergency center) does not have a real-time observation of the environment, hence it can assign the UAVs to the high-priority regions and then UAVs can take into account their real-time observations of the network in order to determine their optimum solution. The model is scalable in the sense that adding or removing the UAVs does not impact the performance of other UAVs as they work independently. To the best of our knowledge, this work is one of the first ones to address the problem of spectrum shortage for UAV networks during critical missions.


The rest of the paper is organized as follows. Section \ref{related} studies the related work regarding spectrum management in UAV networks. Section \ref{sec:System_Model} discusses the system model, assumptions and formulation details. In Section \ref{algorithm}, we propose two search algorithms from the emergency center view, then introduce a multi-agent Q-learning algorithm and analyze its convergence. Section \ref{Simulation} presents the simulation results and discusses the performance of the proposed methods in different settings. Finally, we conclude the study and give some future directions in Section \ref{Conclusion}.

%% file: Texfiles/related.tex
\section{Related Work}\label{related}
While the spectrum scarcity will be a serious challenge in UAV networks given the increasing number of UAVs and the requirements of advanced wireless services, the problem of spectrum management in these networks has been barely investigated so far. 
mmWave communication has been discussed as an option for payload communications of UAVs as a part of 5G, but the technology is not widely available yet. Moreover, the mmWave communication is impacted by high propagation path loss, thereby, it requires the UAVs to be equipped with high directional antennas   to avoid blockage zones and maintain a LOS communication.

The few existing works related to spectrum sharing with drones mainly focused on the coexistence of UAV networks with cellular networks and adopted common notions of spectrum sharing such as interweave method to let the UAVs opportunistically access the spectrum holes of other communication systems, or the underlay method to allow the UAVs to utilize the spectrum of other  systems while maintaining a low interference level \cite{Zhang,DSA_Milcom,Marojevic,Wang,Lyu,Brwon_Policy,Brown_DSA,mousavi2018leader,guan2017smart}. The spectrum sensing method is not an ideal option noting the considerable energy consumption involved in searching a wide range of frequencies. 
More importantly, the spectrum holes are often sparse and appear on different frequencies, therefore they cannot offer continuous communication for the UAV system or require frequent changes of the operating frequency. The spectrum sharing techniques based on databases control (e.g., TV white space) only allow a low level of transmission power for unlicensed users, and allocate a wide and static protection zone around the incumbents. One common drawback of these conventional spectrum sharing methods is that the spectrum owners are oblivious to the presence of the devices seeking for spectrum, but a dynamic and efficient practical spectrum sharing model cannot be implemented unless different users including the spectrum owners and the ones looking for spectrum interactively cooperate with one another.

A primary model for cooperative spectrum sharing in UAV networks where the UAVs are divided into two clusters of sensing and relaying UAVs is proposed in  \cite{shamsoshoara2019distributed}. In this model, the UAVs are assumed to be located in fixed positions with no movement and a multi-agent reinforcement learning-based solution was developed to find the best task allocation for each UAV in a distributed manner.
While this solution can be applicable to scenarios where the communication among the UAVs is not available or reliable, such fully distributed approach for the task allocation with no message exchange among the agents does not guarantee an optimum outcome. 
In \cite{shamsoshoara2019solution}, the problem of cooperative spectrum sharing between a UAV network and a terrestrial network was considered, where a fully centralized approach is proposed serving for both task allocation between the sensing and relaying UAVs, and trajectory optimization. The authors assumed that the emergency center has full control over the UAVs in terms of the task assignment and movement. This problem was solved using a "team Q-learning" algorithm for a multi-UAV scenario. Although the proposed method is able to find the best solution considering the UAVs' locations and assigned tasks; however, it takes a considerable amount of time for the learner to get experience in large grids. Therefore, this centralized model is not scalable and it also relies on a full real-time knowledge of the central controller about the network and the UAVs' status.

In this paper, we propose a hierarchical joint task allocation and path planning method that offers an scalable and reliable solution for cooperative spectrum sharing for UAV networks. In this model, first the emergency center determines the operation region and the proper task of each UAV using a search algorithm or a bipartite graph matching method \cite{ben2008distributed,acevedo2013decentralized}. Then, the UAVs determine their optimum location within their operation region given their real-time observation of their environment. 
The UAV coordination for all agents is performed using Reinforcement Learning (RL) to achieve a common goal. In fact, by using the hybrid approach, the nature of the problem which was a Multi Agent RL is simplified to several single agent RL sub-problems.

%% file: Texfiles/system.tex
\section{System Model}\label{sec:System_Model}
Let us assume a fleet of $N$ autonomous UAVs with the mission of transmitting high data rate surveillance data such as video or images to a pre-defined emergency center. We assume that all UAVs are autonomous in a way that they can control their path in their region and determine their optimal location autonomously. The emergency center can be considered in a form of a ground station or a High Altitude Platform (HAP) in a fixed location. This emergency center has a prior knowledge about the impacted areas and their level of priority, but it may not necessarily have a real-time observation of the operation field.

\begin{figure*}[bht!]
\centering
\includegraphics[width=0.800\paperwidth]{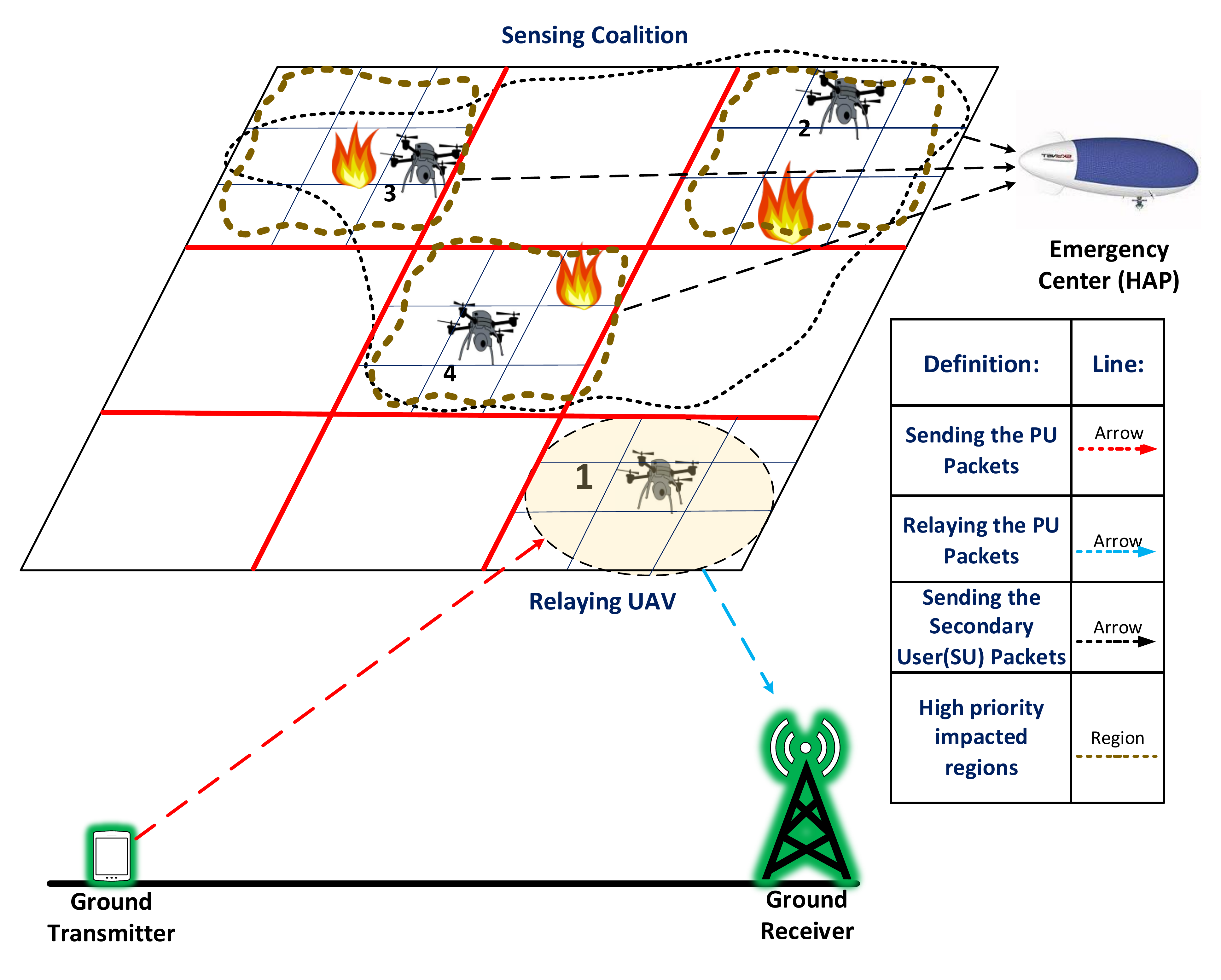}
\caption{A sample scheme of the proposed spectrum sharing between a UAV network and a terrestrial licensed network. Those red lines demonstrate the clustering of the whole area after prioritizing the impacted area and also the size of each region. The regions in the red lines are non-overlapped. 
}
\label{fig:system_grid_1}
\end{figure*}

It is assumed that there exists a licensed pair of terrestrial transmitter-receiver willing to share their spectrum with the UAV network as they suffer from a low quality of communication or they are interested to improve the communication quality through relaying service by the UAVs. For the sake of simplicity, we assume that the primary's transmitter and receiver are too far from each other that the direct communication is not feasible or efficient for them \cite{palat2005cooperative,fan2018optimal,chen2017optimum}. 
Hence, they need a relay to forward their messages for them. We should note that the proposed spectrum sharing method can be also utilized in scenarios where the direct link between the primary's transmitter and receiver exists but the PU takes part in leasing its spectrum in exchange for diversity gain using cooperative relaying methods. The pair of PU's transmitter and receiver are located in a fixed location on the ground and the UAVs (the secondary users) are assumed to hover in a fixed altitude during their mission.

It is assumed that the UAVs are provided with a dedicated bandwidth for Control and Non-Payload Communication (CNPC) to exchange signaling and controlling information and also with a limited bandwidth for payload communications which may not be adequate to deliver real-time transmission of high-resolution data such as video and image during disaster relief operations. Therefore, the UAV network may require additional temporary spectrum access. To meet this demand, one UAV in the network is selected by the emergency center to act as an aerial relay for the terrestrial network, while the rest of the UAVs can take advantage of the spectrum access provided by the primary network to transmit their collected information to the emergency center. The PU grants half of its time of spectrum access to the UAV network and in exchange one selected UAV assists the PU to deliver its packet to the legitimate receiver. In summary, the UAVs can be categorized into two sets of \textit{sensing} and \textit{relaying} UAVs. One UAV is selected as a relay and $N-1$ operate as the sensing UAVs.

We consider a hybrid scheme consisting of both the centralized and autonomous control scenario in which the task of the UAVs is determined by the emergency center while they have the capability to decide for their actions in terms of the mobility to maximize their throughput and lifetime. Thus, the controller does not require continuous updates on the environment status to determine the real-time location of the UAVs, rather it assigns the sensing UAVs to high-priority regions based on its prior knowledge of the environment in the beginning of the mission. Then it revises this allocation as needed if the level of dynamicity such as PU's location or the fire's growth exceeds some pre-defined threshold.




Figure \ref{fig:system_grid_1} demonstrates a sample scenario with three sensing and one relay UAVs.
The UAVs are assumed to be located in a plane (not the emergency center).
The emergency center identifies the high priority impacted areas and clusters these regions to multiple non-overlapping operation fields. The size of operation regions depends on several factors including the application type, the number and type of the UAVs, the shape and dimensions of the impacted areas.
The emergency center assigns the UAVs to the optimal operation field for them based on the residual energy and number of hops they have to fly to reach the intended region and then the UAVs can fly in their operation region and find their optimum location within their region by taking one of the actions of $\{ \textnormal{Up }, \textnormal{Down }, \textnormal{Left }, \textnormal{Right }, \textnormal{Stay}\}$.

In each time slot, the channel state information (CSI) between all emergency nodes in the network including the emergency center and the UAVs
is determined based on the distance between the chosen source and destination followed by a simple LoS model.
The CSI parameters between the PU and the relaying UAV follow the slow
Rayleigh fading and are available at the transmitter based on common channel estimation techniques \cite{sun2002estimation,coleri2002channel,Afghah_INFOCOM}.

In this work, we defined the CSI parameters as follow: $h_{PT, U_i}$ expresses the CSI between the primary transmitter and $i^{th}$ UAV, $h_{U_i, PR}$ describes the channel between the $i^{th}$ UAV and the primary receiver, $h_{U_i, E}$ carry the information for channel between $i^{th}$ UAV and the emergency center.
The noise of all channels are modeled at the receiver side with normally distributed symmetric complex values, $Z \sim CN(0, \sigma^2)$. We assume that all nodes have a constant arbitrary value for the transmission power. These arbitrary values are different for the UAVs, the source, and the primary transmitter. While several works such as \cite{shamsoshoara2015enhanced, shah2018deep, mohammadi2018qoe} focused on optimizing the power consumption, we consider a constant transmission power; however, we address the energy consumption rate by minimizing the number of movements in the plane and each region.



\begin{figure}[t]
\centering
\includegraphics[width=0.5\linewidth]{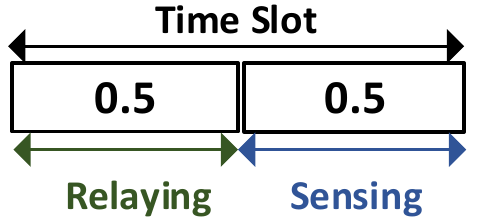}
\caption{The allocation of each time slot between the relaying and sensing tasks}
\label{fig:timeslot}
\end{figure}

We consider half-duplex transmission for the primary network.
In the first half of the time slot, the primary transmitter sends its packet to the relaying UAV and then the UAV forwards the packet to the primary receiver. The sensing UAVs transmit the gathered information such as video or sensed data to the emergency center during the second half of the time slot. Figure \ref{fig:timeslot} illustrates the time allocation between two networks in each slot.

Different relaying schemes such as amplify and forward (AF) and decode and forward (DF) can be utilized at the relay UAV \cite{souryal2006performance,levin2012amplify,guan2016distributed,chen2015relay}. Here, we utilize the AF scheme, hence the throughput rate for the primary user can be achieved as \cite{laneman2004cooperative}: 
\begin{align}\label{eq:R_PU}
R_{PU} & = \frac{1}{2}\log_2(1 + P_{PT}|h_{PT,PR}|^2  
\\ \nonumber
&
+ \frac{P_{PT}|h_{PT,U_i}|^2 \: P_{U_i}|h_{U_i, PR}|^2}{\sigma^2 +P_{PT}|h_{PT,U_i}|^2 + P_{U_i}|h_{U_i, PR}|^2}),
\end{align}
where the transmission power for the primary transmitter and UAV $i$ are denoted by $P_{PT}$ and $P_{U_i}$.  Background noise power at the receiver is $\sigma^2$, and let $i$ denote the index for the $i^{th}$ (UAV).
 Due to the long distance between
 the PT and PR, $h_{PT, PR}$ is zero, hence we can eliminate this term from the throughput rate ($|h_{PT, PR}| \simeq 0 $).

The throughput for each sensing UAV can be calculated as \cite{yu2005amplify,nosratinia2004cooperative}:
\begin{align}\label{eq:R_SE}
R_{SE_i} &  
= \frac{\lambda_i}{2}\log_2(1 + \frac{P_{U_i}|h_{U_i, E}|^2}{\sigma^2}),
\end{align}
where $\lambda_i$ is the portion of time that is allocated to $i^{\text{th}}$ UAV transmission considering the priority of its operation field.

The achievable rate at the emergency center, $R_{SE}(\textnormal{Multi-UAV})$ can be obtained as
\begin{align}\label{eq:R_SE_multi}
R_{SE} & \textnormal{(Multi-UAV)}  = \sum_{i \in U_E} R_{SE_i}
\\ \nonumber &
= \sum_{i \in U_E} \frac{\lambda_i}{2}\log_2(1 + \frac{P_{U_i}|h_{U_i, E}|^2}{\sigma^2}),
\end{align}
where $U_E = \{E_1, E_2, \dots E_{N-1} \}$ denotes the sensing set. 

\begin{figure}[bht]
\centering
\includegraphics[width=1\linewidth]{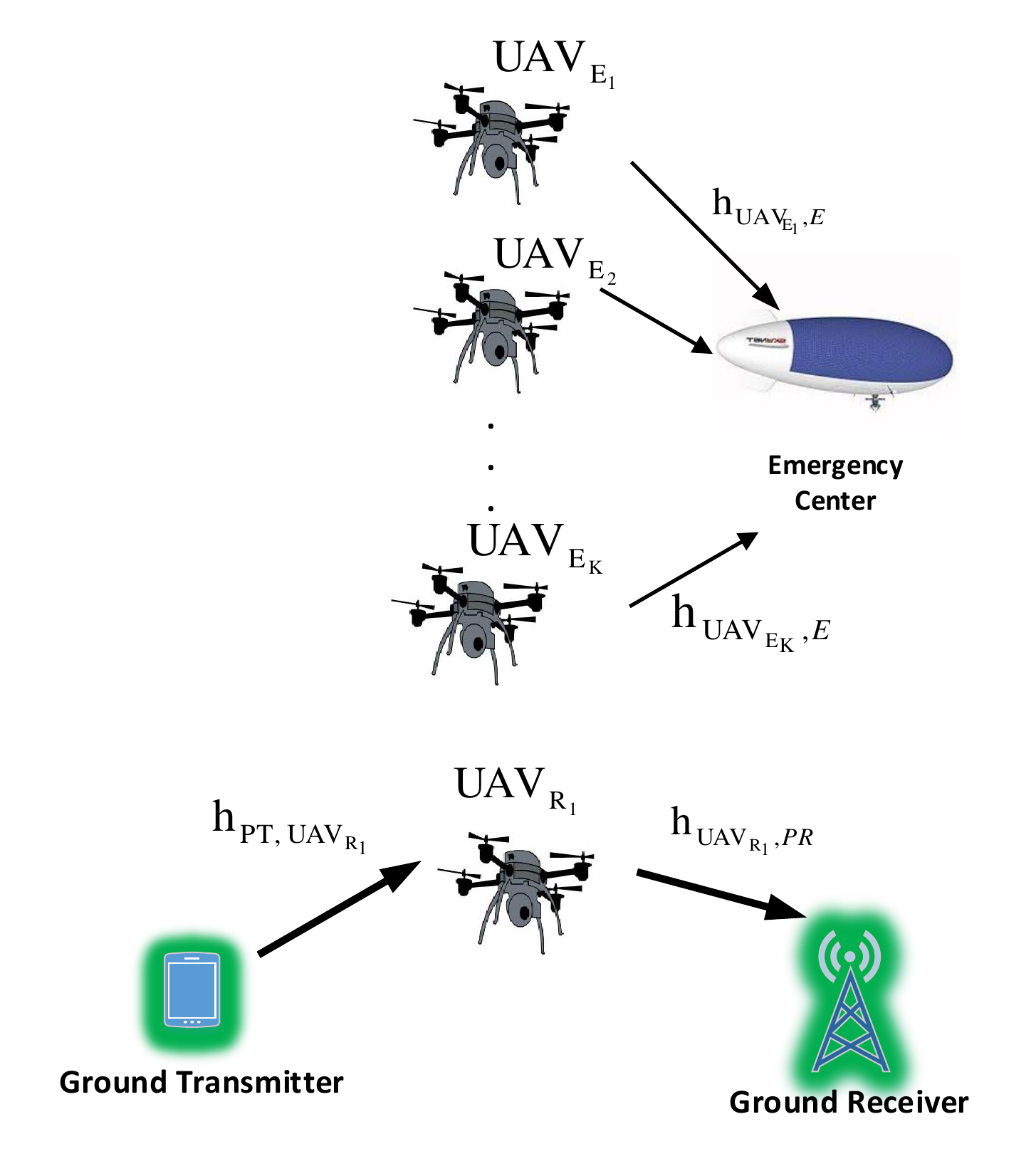}
\caption{Channel state information for the relaying UAV and sensing coalition.}
\label{fig:multiUAV_system}
\end{figure}

In Figure \ref{fig:multiUAV_system}, a sample case with $(N-1)$ sensing UAVs and one relaying UAV  is shown. Figure \ref{fig:multiUAV_system} shows channel state information between different nodes in this scenario.

%% file: Texfiles/algorithm.tex
\section{Proposed Algorithm}
\label{algorithm}
The proposed algorithm for spectrum sharing includes two steps: (i) the search algorithm which picks the best region for the primary network based on the throughput metric and assigns UAVs to the prioritized regions and the primary region to prolong the network lifetime, and (ii) the reinforcement learning algorithm performed by the individual UAVs to gain experience on the states and actions to decide movement actions wisely and maximize the network throughput and lifetime.  

\subsection{Task Allocation and Region Assignment by the Emergency Center}
\label{subsec:Search}
We assume that at the beginning of each round of optimization, the UAVs are located at random cells across the region with different random initial energy. The emergency center determines the number of regions in the grid-operation field knowing the number and type of available UAVs. It also takes into account its prior knowledge of the impacted field to identify the high priority regions. After determining the high-priority regions and their IDs, the emergency center identifies a proper primary network within the operation field and assigns the best relay UAV to the region close to the PU in order to maximize the throughput rate of the PU. 
Selecting the region for the relaying UAV depends on the Euclidean distance between the region and the primary receiver using the search algorithm described in (\ref{eq:searchprimary}). 
\begin{align}\label{eq:searchprimary}
\textnormal{UAV}_P = \argmax_{u \in \mathcal{U}}  f(u) := \{u | u \in \mathcal{U} : E(u) - d(u) \Psi \},
\end{align}
where, $\textnormal{UAV}_P$ is the index of the chosen UAV to serve the primary network, $\mathcal{U}$ is the set of all UAVs, $E(u)$ is the initial energy for UAV $u$, $d(u)$ is the number of steps or hops that UAV $u$ should travel to reach the center of the primary region, and $\Psi$ is the energy consumption rate per movement. Finding the $u$ guarantees the longest lifetime for the primary network. After assigning the primary UAV to the primary region, the UAV's initial energy will be decreased due to the flight distance. 

Next, the emergency center chooses arbitrary number of regions
based on the disaster operation and the priorities for the sensing area. To utilize all UAVs, we assume that the arbitrary number of regions is equal to number of sensing UAVs. Afterward, it assigns the remaining UAVs to the high-priority regions such that the residual energy after the flight distance from the initial location to the region is maximized. 
The emergency center makes a preference list including the sorted indices of the regions for each UAV. Next, it allocates the UAVs to the regions based on the preference value. If two or more UAVs have the same preference value for one specific region, then the emergency center assigns the UAV with higher remaining energy to this region and allocates the second UAV to its next preference region. Expression (\ref{eq:searchsecondary}) shows the preference array for the sensing UAV $i$: 
\begin{align}\label{eq:searchsecondary}
\textnormal{R}_{S_i} = \argsort_{r \in \mathcal{R}}  g(r) := \{r | r \in \mathcal{R} : E_i - d_i(r) \Psi \},
\end{align}
where, $\argsort\limits_{r \in \mathcal{R}} g(r)$ is the function to order a list in a descending order and returns the index of those values, $R_{S_i}$ is the preferred regions for UAV $i$, $\mathcal{R}$ is the set of high-priority regions for the secondary network. $E_i$ is the initial energy for UAV $i$ at its first location. $d_i(r)$ is the distance of UAV $i$ from its initial location to the region $r$. 

\begin{figure*}[bt]
	\centering
	\begin{subfigure}[bht]{0.45\textwidth}
         \centering
         \includegraphics[width=\textwidth]{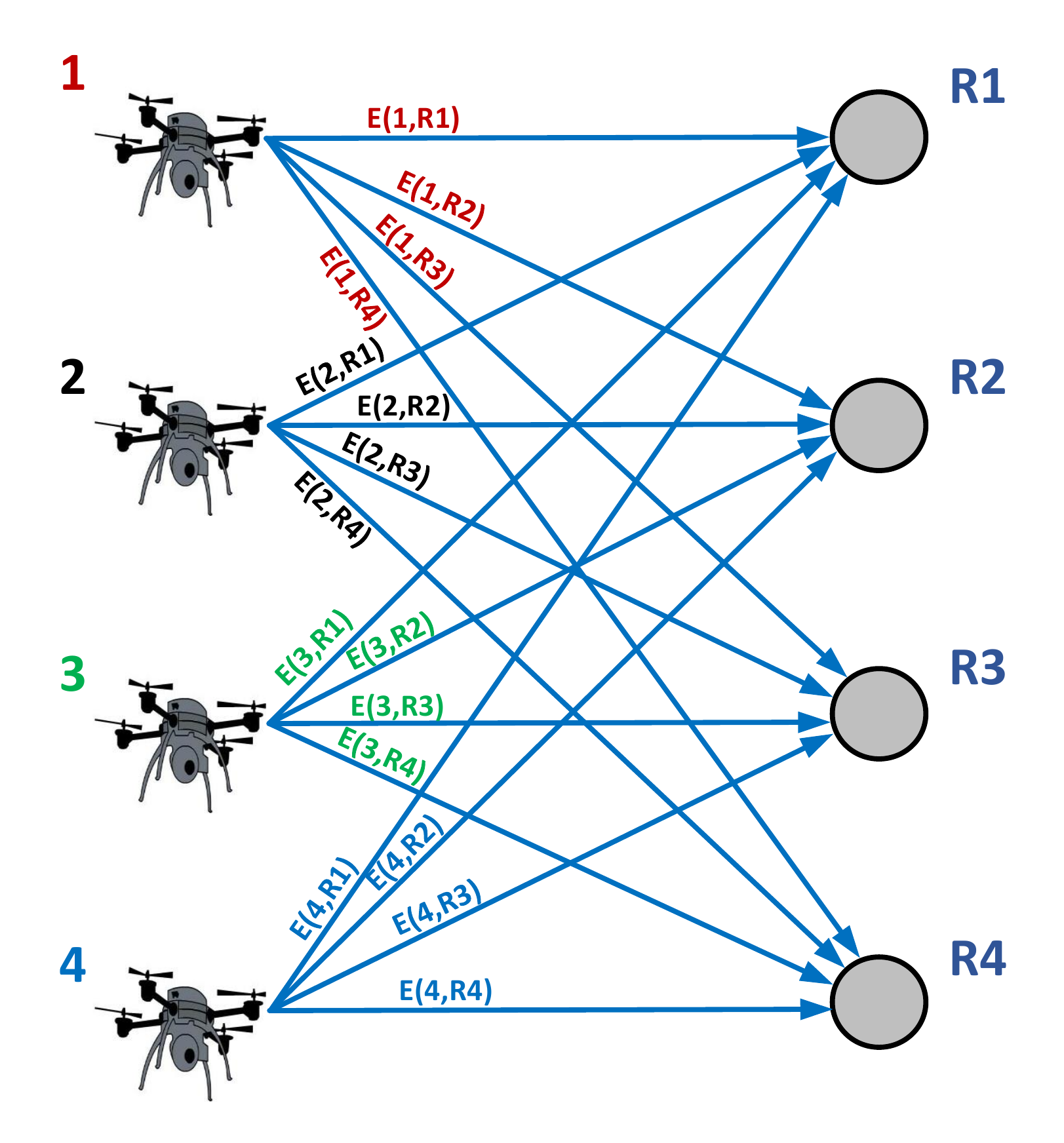}
         \caption{A sample complete graph for region allocation}
         \label{subfig:completeGraph}
     \end{subfigure}
     \hfill
     \begin{subfigure}[bht]{0.45\textwidth}
         \centering
         \includegraphics[width=\textwidth]{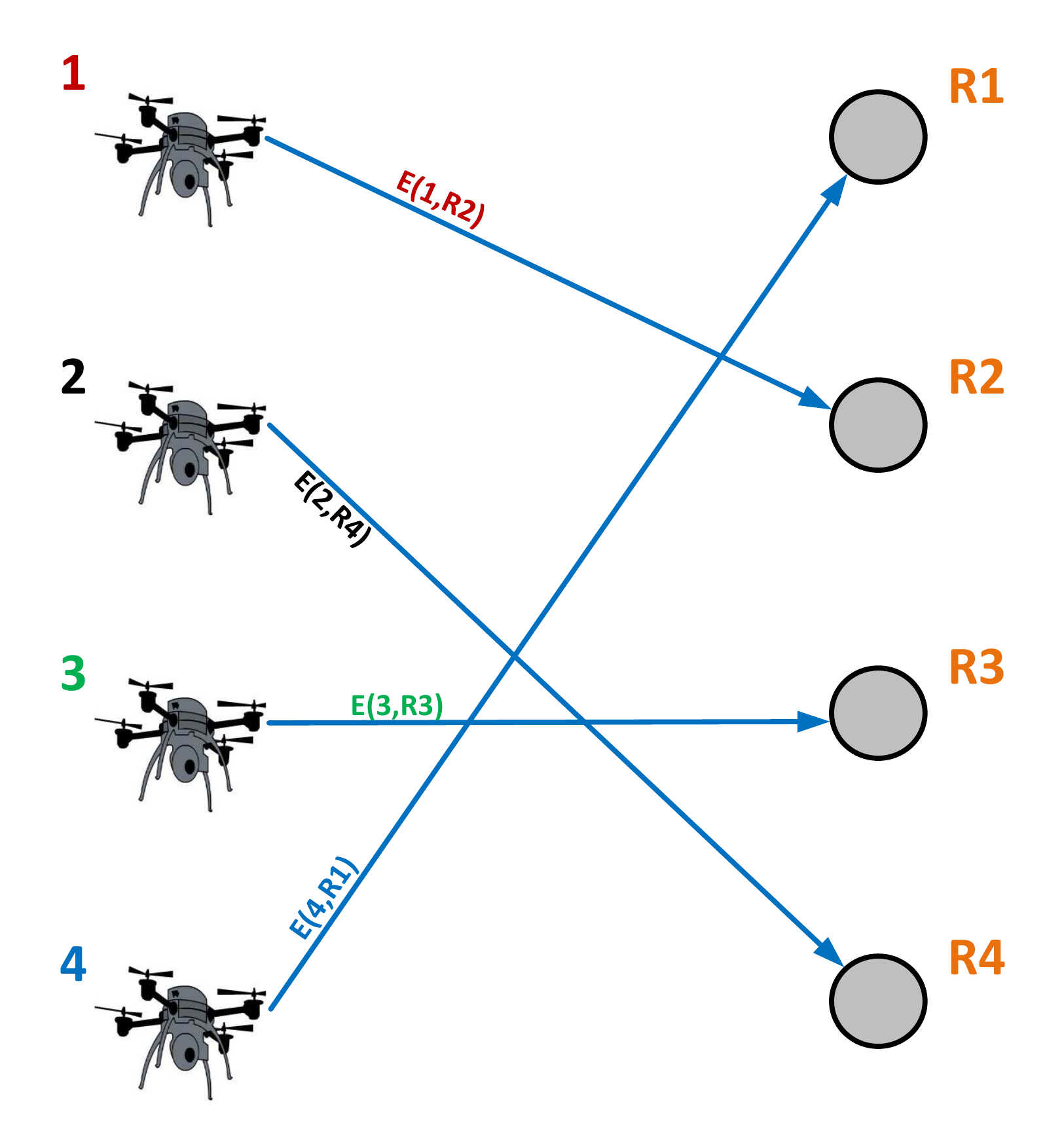}
         \caption{An sample UAV allocation to the high-priority regions}
         \label{subfig:graphmatch}
     \end{subfigure}
     \caption{An example for matching the UAVs to the high priority regions }
    \label{fig:secondarynetwork}
\end{figure*}

Figure \ref{subfig:completeGraph} shows a complete graph with the sensing UAVs and the region centers as the vertices and the remaining energy after the flight distance as the edges.  Figure \ref{subfig:graphmatch} shows an example of the allocated UAVs to the regions based on the preference regions in (\ref{eq:searchsecondary}). This allocation mechanism guarantees the longest lifetime for each UAV. Based on this ranking criteria, the goal of region selection is to have an optimal throughput rate and the purpose of the UAV allocation is to get the longest lifetime for both networks considering the energy.
After allocating the UAVs to the regions, each UAV utilizes a Q-learning algorithm to find the best cell in each region. 

\subsection{Location Optimization at Individual UAVs using reinforcement learning}\label{subsec:RL}
In applications where training data-sets are available to learn the optimal behavior, the artificial neural network-based (ANN) approaches are commonly utilize to learn the optimal solution. However, in applications such as our problem that lack the required training data-sets or knowledge about the optimal solution, reinforcement learning-based methods can offer a trial-and-error approach to determine the solution. ANN-based methods are limited to the given training data-sets for the optimal solution and the obtained solution cannot be better than the provided samples. However, a key advantage of the RL-based approaches is that the agent can learn to behave better than an expert in the problem.

In this study, we break down the Multi-Agent Reinforcement Learning (MARL) into several single RL sub-problems. Each UAV is considered as a single agent which is independent of other UAVs. Each UAV is connected to its region with action and perception \cite{kaelbling1996reinforcement, mousavi2017traffic,mousavi2016deep}.


Based on the aforementioned system model in Section \ref{sec:System_Model}, each UAV operates in an individual region area with no interaction with other UAVs. The states, actions, and rewards are exclusive for each UAV. Taking an action for a specific UAV does not impact the state and reward for other UAVs. As a result, a simple Q-learning is used to find the best location for each UAV. Each UAV monitors its behavior and decides based on its experience.

The decision making for each UAV is modeled by a Finite Markov Decision Process (FMDP) framework to construct the decision making in the discrete time stochastic control process. Each UAV only observes its own state refers to its location within its allocated region. Assume that $S_t^{i} \in \mathcal{S}^i$ defines the state for the $i^{\text{th}}$ UAV at step t, $\mathcal{S}^i$ is the set of all states for UAV $i$. Taking an action such as $a_t^{i} \in \mathcal{A}^i$ for $i^{th}$ UAV changes its state from $S_t^{i}$ to $S_{t+1}^i$. Choosing an action for each UAV defines its behavior which follows the policy $\pi(S)$ based on (\ref{eq:policy}).
\begin{align}\label{eq:policy}
\pi(S_t) = \argmax_{a_t \in \mathcal{A(S}_t\mathcal{)}} \pi(S_t, a_t),
\end{align}

The default action set $\mathcal{A}$ includes four actions for movement and one action for staying at the previous location $\mathcal{A} = \{  \uparrow(0), \downarrow(1), \leftarrow(2), \rightarrow(3), -(4) \}$. Depending on the UAV's location, some particular actions may be prohibited. For instance, if the UAV is located on the region's edge side, taking an action which results to leave the region is prohibited. Taking a new action and altering into a new state ($S_{t+1}$) updates the reward value $r_t \in \mathcal{R}$ in respect to (\ref{eq:reward}). Following the optimal policy based on (\ref{eq:policy}) guarantees the action set which results into expected reward maximization $ \mathbb{E}(\sum\limits_{t=0}^{\infty} \gamma^{t} r_{t})$, where $\gamma \in [0, 1)$ is the discount factor. Individual UAV forms distinct MDP including a 4-tuple which can be shown as $(\mathcal{S}, \mathcal{A}, P_a, r_a)$, where $\mathcal{S}$ is finite set of states for each UAV. Since all the regions sizes are the same and the UAVs are independent, all UAVs consider the same state set, $\mathcal{A}$, a finite action set which depends on the UAV's location in the region. $P_a: S_t \times A \times S_{t+1} \rightarrow [0, 1]$ is the state probability function which stands for the randomness in transition between the $S_t$ and $S_{t+1}$, $r_a: S \times A \times S_{t+1} \rightarrow \mathbb{R}$ is the gained scalar reward for an individual UAV if it takes an action $a$ and changes its state from $S_t$ to $S_{t+1}$. This reward introduces the effect of the immediate action. However, it does not interpret the long-term effect of the action. Hence, the UAV needs to maximize the long-term reward. One approach is using the optimal action-value function and computing that with the Q-learning algorithm \cite{mousavi2017applying, mousavi2016learning,mousavi2018researching} which satisfies the Bellman optimality equation. All UAVs update their Q-values based on:
\begin{align}\label{eq:Qupdate}
Q(S_t, a_t) = & (1-\alpha)\times Q(S_t, a_t)
\\ \nonumber & 
+ \alpha [r_a(S_t, S_{t+1}) + \gamma \max_{a' \in \mathcal{A}} Q(S_{t+1}, a')],
\end{align}
where, $\alpha \in (0, 1]$ is the learning rate for the agent. $Q(S_t, a_t)$ is the action-value for the current state which depends on the $\alpha$, current reward, $\gamma$, and the $\max_{a' \in \mathcal{A}} Q(S_{t+1}, a')$. Last term stands for the maximum action-value among all values for the next state $S_{t+1}$. While (\ref{eq:Qupdate}) defines the Q table and values for one UAV, (\ref{eq:multiQupdate}) introduces the updating matrix for all agents:

\begin{flalign}\label{eq:multiQupdate}
  & \begin{bmatrix} 
    Q_1(S_t^{(1)}, a_t^{(1)}) \\ Q_2(S_t^{(2)}, a_t^{(2)}) \\ \vdots \\ 
    Q_N(S_t^{(N)}, a_t^{(N)}) 
 \end{bmatrix}
 =
\end{flalign}

 \begin{flalign}
   \nonumber
 & 
 \qquad
  \begin{bmatrix}
   1-\alpha^{(1)} \\ 1-\alpha^{(2)} \\ \vdots \\ 1-\alpha^{(N)}
  \end{bmatrix}
  \cdot
  \begin{bmatrix} 
    Q_1(S_t^{(1)}, a_t^{(1)}) \\ Q_2(S_t^{(2)}, a_t^{(2)}) \\ \vdots \\ 
    Q_N(S_t^{(N)}, a_t^{(N)})
 \end{bmatrix}
  + 
  \begin{bmatrix}
   \alpha^{(1)} \\ \alpha^{(2)} \\ \vdots \\ \alpha^{(N)}
  \end{bmatrix}
  \\
 \nonumber
 & 
  \cdot
  \begin{bmatrix} 
    r_a^{(1)}(S_t^{(1)}, S_{t+1}^{(1)}) + \gamma \max\limits_{a'^{(1)} \in \mathcal{A}^{(1)}} Q(S_{t+1}^{(1)},a'^{(1)}) \\
    r_a^{(2)}(S_t^{(2)}, S_{t+1}^{(2)}) + \gamma\max\limits_{a'^{(2)} \in \mathcal{A}^{(2)}} Q(S_{t+1}^{(2)}, a'^{(2)}) \\
    \vdots \\ 
    r_a^{(N)}(S_t^{(N)}, S_{t+1}^{(N)}) + \gamma\max\limits_{a'^{(N)} \in \mathcal{A}^{(N)}} Q(S_{t+1}^{(N)}, a'^{(N)})
 \end{bmatrix}, 
\end{flalign}
where $N$ is the number of agents (UAVs), also we assumed that all UAVs have the same learning rate and discount factor. The Q-tables are distinct for each UAV and the UAVs take actions simultaneously.

To define the reward function for the agents based on their actions in the previous time slots, we assume that the both the primary receiver and emergency center report their received throughput to the UAVs.
Since the goal of our model is to maximize the throughput, network lifetime, and minimize the energy consumption, the reward function consists of both the gained throughput and remaining battery of the node. The reward function compares the last two consecutive throughput rates and remaining energy to give an award to UAVs. Since the UAVs are located in different regions, their reward functions are independent of each other and there is no correlation between them. (\ref{eq:reward}) defines the reward function for each UAV:



\begin{align}\label{eq:reward}
&Reward(i) = 
\\
\nonumber
&
\begin{cases}
\beta_1 \quad \textnormal{if } (R(t) > R(t-1)) \textnormal{ and } (E(t-1)-E(t) = \Psi)
\\
\beta_2 \quad \textnormal{if } (R(t) = R(t-1)) \textnormal{ and } (E(t-1)-E(t) = \psi)
\\
\beta_3 \quad \textnormal{if } (R(t) < R(t-1)) \textnormal{ and } (E(t-1)-E(t) = \Psi)
\end{cases},
\end{align}
where, $R(t)$ is the throughput rate at timeslot $t$,  $\beta_{1}$, $\beta_{2}$, and $\beta_{3}$ are rewards based on the utility and remaining energy values.
$i$ denotes the index for $i^{th}$ UAV. $\Psi$ is the energy consumption rate when a UAV changes its location and performs data transmission. $\psi$ is the energy consumption rate when a UAV stays at its location and performs data transmission. We also assume that the energy consumption due to the UAV's mobility is more than the energy used for data transmission, as it is usually the case in UAV networks.   The reward is obtained based on the throughput rate and since the throughput depends on the distance between any source and destination, the UAVs' goals are to find a proper location for themselves in each region to maximize their throughput
and minimize the energy consumption.

In each step, the UAVs are awarded with different reward values based on (\ref{eq:reward}) which follows three possible options: first, the UAV changes its position and it improves the throughput rate too. But it costs energy to change the cell in the region. In this case, the UAV earns $\beta_1$ as a positive reward. Second, the UAV stays at its position and keeps its previous throughput rate and saves its energy, in this case, the UAV is granted with $\beta_2$ as the reward. Third, the UAV changes its location and receives less throughput rate and it also looses its energy because of the movement. In this case, the UAV is punished with $\beta_3$ as a negative value.

For the action selection process, 
we implement the $\epsilon$-greedy exploration for all UAVs with the constant $\epsilon$. Hence, each UAV chooses a random action with the probability of $\epsilon$ and it chooses the best action with the probability of $1-\epsilon$ to find the best action based on updated values in the Q-table. We like to note that in team Q-learning, a single learner decides for all agents in a random or a greedy manner. However, in multi-agents such as Nash Q-learning or the method in this paper, the UAVs act independently. For instance, some UAVs behave randomly while others choose greedy actions \cite{sutton2018reinforcement}. The utilized action selection method is shown in \ref{eq:epsgreedy}: 
\begin{align}\label{eq:epsgreedy}
& a_t^{(i)}:
\begin{cases}
\textnormal{random Actions,} \qquad \ \ \ \
\textnormal{rand(i)} < \epsilon,
\\
\argmax\limits_{a^{(i)}} Q_i(S_t^{(i)}, \mathcal{A}), \qquad  \textnormal{o/w}, 
\end{cases}
\end{align}
where $i \in \{1, 2, \dots, N \}$ is the index for each UAV and $\mathcal{A}$ is the available actions set for the $i^{th}$ UAV when it is located in state $S_t^{(i)}$.

\begin{figure*}[bt]
	\centering
	\includegraphics[width=0.95\linewidth]{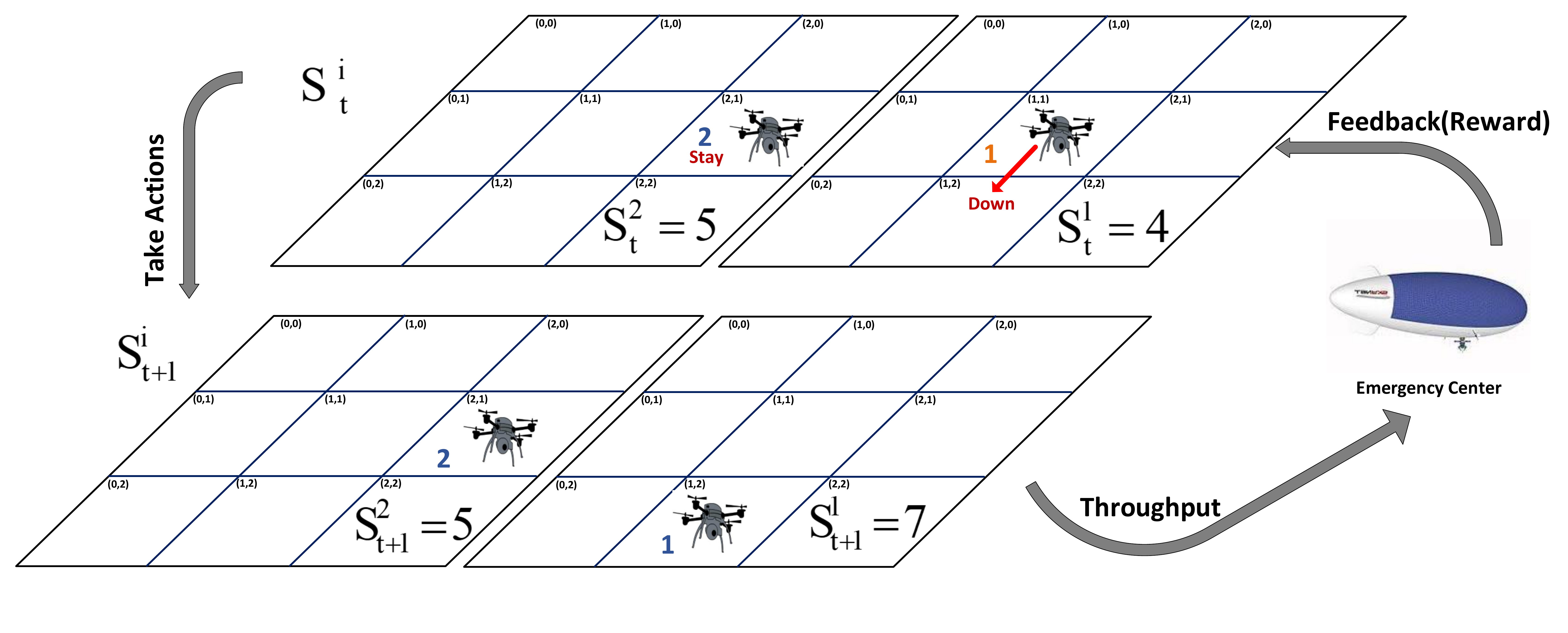}
	\caption{An example of RL-based decision making process for 3 UAVs in a 3$\times$3 grid.}
    \label{fig:state_action_reward}
\end{figure*}

Figure \ref{fig:state_action_reward} demonstrates an example of the learning procedure for two UAVs based on the feedback reward, the taken actions, and the states. UAV $1$ changes its location based on the taken action. 
UAV $2$ stays at its location. These processes bring a new state for UAV $1$, based on the \textit{Down} action, it flies from state 4 to state 7. Also, UAV $2$ keeps its previous state at location 5. Based on the received throughput for both the primary receiver and the emergency center as feedback, the rewards are assigned to the UAVs. 
Algorithm \ref{algo:QMARL} summarizes the search algorithm to find the best primary region and the UAV allocation by the emergency center, also it abstracts the requiring steps for the Q-learning in the RL process for all UAVs. 
\begin{algorithm}[hbtp]
\SetAlgoLined
 \textbf{initialization:}\\
 Set initial parameters and conditions\\
 Select the primary region by the emergency center\\
 Allocate the operation regions for sensing UAVs by the emergency center noting the region priorities\\
 Decrease the energy after UAV allocation\\
Set all initial parameters for the RL
 
 \For{all Runs}
 {
    Initial all Q-Tables, utility functions, actions, tasks, and all arrays to zero\\
    \For{all Episodes}
    {
        Set the location for the UAVs to the initial\\
        Set the energy for the batteries to the initial energy\\
        \If{Variations in the environment (dynamicity level) $>$ $\tau$}
        {
            re-initiate the controller process for the UAVs and regions   
        }
        \For{all $t < Step \ Size$ \textnormal{in the Grid}}
        {
            \For{all the UAVs}
            {
                Get the State from the updated location\\
                \eIf{Random $<$ $\epsilon$}
                {
                    Choose a random action\\
                    Update the location\\
                }
                {
                    Choose an action based on best Q-value from the Q-table\\
                    Update the location\\
                }
                Update the energy after flight\\
                Calculate the utilities and throughput\\
                Update the energy after transmission\\
                Calculate the reward\\
                Update the Q-Table
            }
        }
    }
 }
 \caption{Region assignment, UAV allocation, and Q-learning for RL framework 
 }
 \label{algo:QMARL}
\end{algorithm}

\subsection{Convergence analysis of the RL algorithm}
Since the UAVs are independent, they have different Q-tables.
Let $Q_{i}^*(S, a_{i})$ denote the optimal Q-value for the $i^{\text{th}}$ UAV in algorithm \ref{algo:QMARL}. 

\textbf{Theorem 1}: The utility functions for the system which are defined by formulas (\ref{eq:R_PU}), (\ref{eq:R_SE_multi}), and (\ref{eq:reward}) are bounded and finite.

Proof: Considering (\ref{eq:R_SE_multi}), terms $|h_{S, U_j}|^2$ and $|h_{U_j, E}|^2$ depend on the location of UAVs. Since, the emergency center and the primary users are placed in a fix location and the gird size is bounded for definite positions, then $R_{SE}(\textnormal{Multi-UAV})$ is bounded for definite numbers. The same approach can be used for $R_{PU}$. As a result, both (\ref{eq:R_PU}) and (\ref{eq:R_SE_multi}) are bounded. Also, the reward function is defined based on $\beta_1$, $\beta_2$, and $\beta_3$ which are finite values.

\textbf{Theorem 2}: If the learning rate is bounded between 0 and 1 ($0 < \alpha \leq 1$) for all $(S, A) \in \mathcal{S} \times \mathcal{A}$ which requires all states-actions to be observed for infinite times, then:

\begin{align}\label{eq:learningrate}
\sum_{i=1}^{\infty}\alpha = \infty, \qquad \sum_{i=1}^{\infty}\alpha^2 < \infty
\end{align}

Proof: To prove Theorem 2, auxiliary results from stochastic approximation are needed as provided in \cite{melo2001convergence, jaakkola1994convergence}. 

\textbf{Lemma 1}: The Q-learning algorithm in algorithm \ref{algo:QMARL} with the update rule from (\ref{eq:multiQupdate}) is converging to the optimal $Q_{i}^*(S, a_{i})$ with probability one (w.p.1) if the utility function for the system is bounded, the state and actions sets are finite, and $\sum\limits_{i=1}^{\infty}\alpha = \infty, \  \sum\limits_{i=1}^{\infty}\alpha^2 < \infty$ for $0 < \alpha \leq 1$ \cite{dayan1992q,melo2001convergence}:

\begin{align}\label{eq:Qconvergance}
\lim_{t \rightarrow \infty} Q_{t, i}(S, A) = Q_{i}^*(S, a_i) 
\end{align}
where, $t$ is the time step in each episode.

%% file: Texfiles/simulation.tex
\section{Simulation Results}\label{Simulation}
In this section, we consider different scenarios for different gird sizes and system conditions to evaluate the performance of our proposed method. First, a pair of ground-based transmitter-receiver as the primary users and an aerial-based  emergency center (e.g. an HAP) are located in random locations. The whole area of coverage is surfaced into $L_1 \times L_2$ tiles. The drones are assumed to be located in arbitrary locations at the beginning of the optimization round. They can move in four directions of $\{ \textnormal{Up }, \textnormal{Down }, \textnormal{Left }, \textnormal{Right} \}$ or stay in the same location. The whole grid size is divided into a pre-defined number of regions.  
$h_{i,j}$ defined the CSI parameters between nodes $i$ and $j$ is calculated based on a LoS model with a known propagation loss factor. We assume that the loss factor in our scenario is $-2$. LoS model considers the 3 dimensional Euclidean distance between the nodes. $\beta_1$, $\beta_2$, and $\beta_3$ (the rewards in (\ref{eq:reward})) are chosen arbitrary based on the experiment in the simulations. Random values between 4000 and 5000 Jules are chosen as the initial battery value for the UAVs. The transmission powers are chosen based on $P_{PT}=10~mW$ and $P_{U_i}$=$20~mW$. The mobility consumption rate ($\Psi$) and the transmission consumption rate ($\psi$) in each time slot are $10.0 \frac{\textnormal{J}}{\textnormal{Mobility}}$ and $0.5 \frac{\textnormal{J}}{\textnormal{transmission}}$, respectively. $\sigma^2 = 1nW$. RL parameters $\alpha$, $\gamma$, and $\epsilon$ are chosen as 0.1, 0.3, and 0.1, respectively unless mentioned explicitly otherwise. For the RL algorithm, we assume that the rate of the exploration-exploitation is fixed meaning that the UAVs with a constant probability ($\epsilon$) choose the actions based on the best action-state Q-values otherwise they do it randomly.

Since each UAV is located in a separate region, the state-space is $(R_1 \times R_2)$, which $R_1$ and $R_2$ are the dimensions for the region area. All regions have the same dimension during the simulation unless the emergency center changes them at the beginning of each run. The number of UAVs is 5 in all iterations and episodes. The emergency center applies a simple search algorithm to find the best primary region and the relay UAV. Also, it uses the bipartite graph matching to allocate UAVs to the regions. The time complexity to find the best primary region is $O(M)$, where $M$ is the number of regions and the time complexity for finding the best relay UAV is $O(N)$, where $N$ is the number of UAVs. Next, the emergency center assigns the sensing UAVs to the high priority-regions with the time complexity of $O(N)$ and utilizes the bipartite graph matching. It matches the UAVs to the high-priority regions with the time complexity of $O((N-1)^2) \sim O(N^2)$.

20 external iterations are considered to improve the accuracy of the simulation and avoid intolerance of the initial values and a biased behavior. In each iteration, the UAVs run 40 episodes to fill the gaps of Q-values in their Q-tables. The Q-tables are initialized with zeros and the final Q-values in each Q-tables is the initialization for the next episode. The number of steps in each episode is assigned based on the experiment and the number of states in each region. Table \ref{tab:sim} brings the parameters and timing values for all the executed simulations. Timing values are calculated using a system with AMD RYZEN 9 3900X CPU @ 3.8GHz and 64.0GB RAM @ 3200MHz. Measured timing values are dedicated to each episode and then summed up over all episodes and finally over all iterations. Compared to the methods such as team Q-learning which require one single learner with a huge Q-table and state-action values, this method that employs multi learners with no correlation has much better time complexity.

\begin{figure*}[bt]
	\centering
	\begin{subfigure}[b]{0.25\textwidth}
         \centering
         \includegraphics[width=\textwidth]{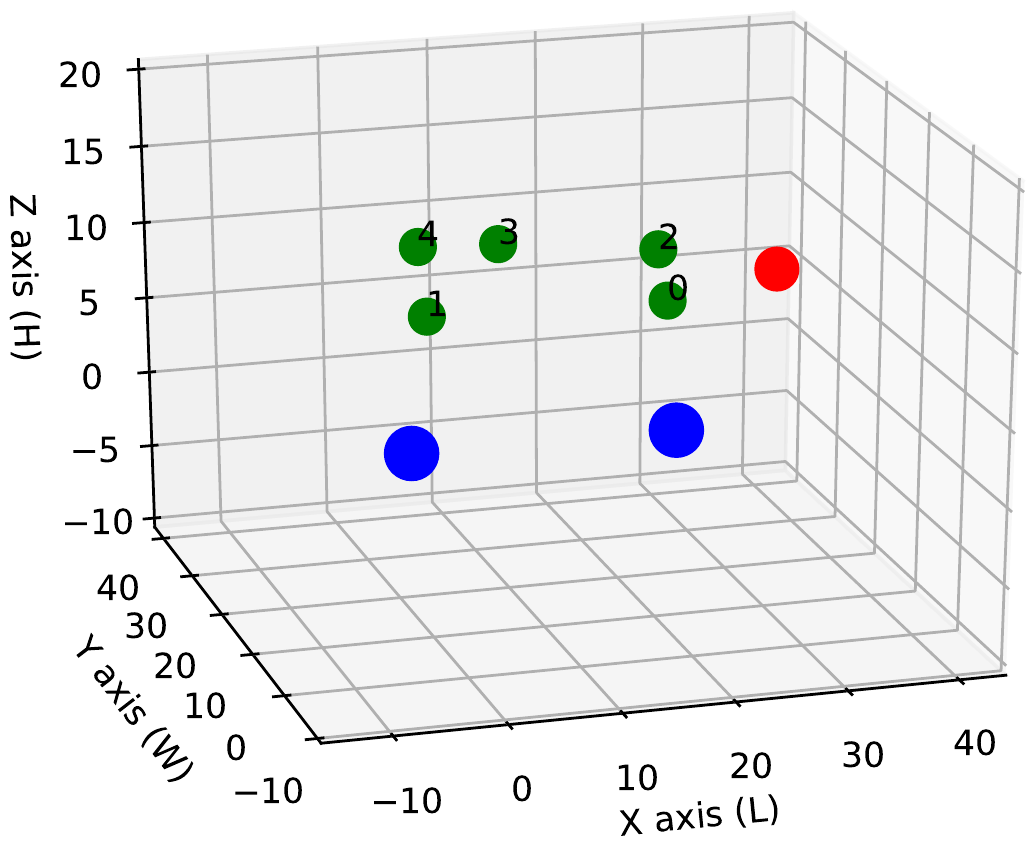}
         \caption{3D topology before regions assignment}
         \label{subfig:3dtopo}
     \end{subfigure}
     \hfill
     \begin{subfigure}[b]{0.35\textwidth}
         \centering
         \includegraphics[width=\textwidth]{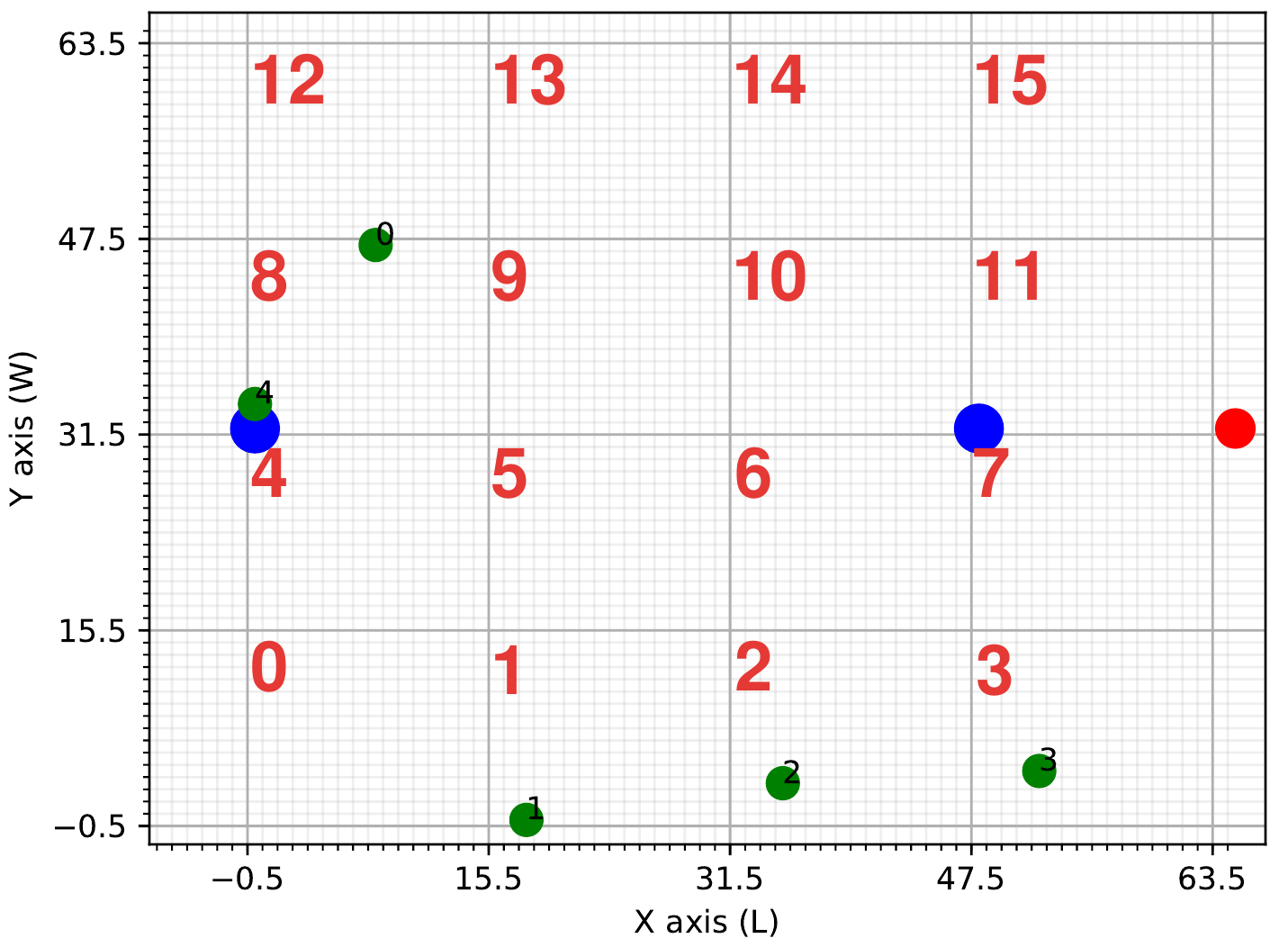}
         \caption{2D topology before regions assignment}
         \label{subfig:2dtopo_before}
     \end{subfigure}
     \hfill
     \begin{subfigure}[b]{0.35\textwidth}
         \centering
         \includegraphics[width=\textwidth]{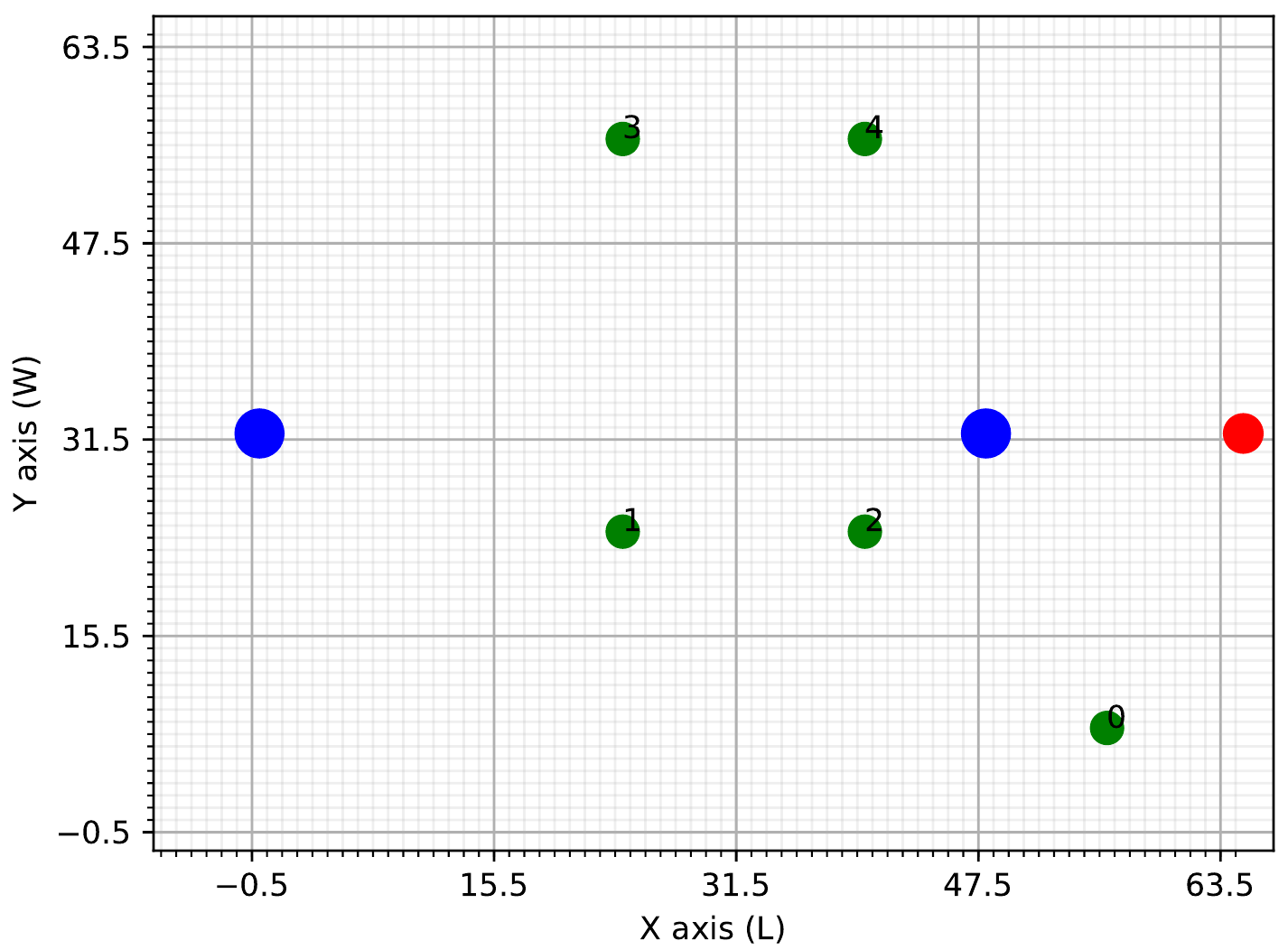}
         \caption{2D topology after regions assignment}
         \label{subfig:2dtoporl}
     \end{subfigure}
    \caption{Topology in 2D and 3D for 5 UAVs in a 64 $\times$ 64 grid size and 16 regions}
    \label{fig:topology}
\end{figure*}

The initial random locations for 5 UAVs and fixed locations for the primary users and the emergency center in a constant altitude are shown in Figure \ref{fig:topology}. Figure \ref{subfig:3dtopo} shows the 3 dimensional topology for all UAVs with a random initialization, and Figure \ref{subfig:2dtopo_before} demonstrates the initial point but in a $Z$ plane. Finally, Figure \ref{subfig:2dtoporl} shows the 2D topology for the UAV after the emergency center made decision about the region assignments. 
Red, blue, and green nodes specify the emergency, primary, and UAV users, respectively.

\begin{table*}[htbp]
\caption{Simulation parameters and required times for the simulation with 2 UAVs and 6 actions.}
 \centering{
\label{tab:sim}
    \resizebox{1\linewidth}{!}{  
\begin{tabular}{c|cccccccccc}
\toprule
\toprule
\textbf{Grid Size:} & {9$\times$9 (81)} & {16$\times$16 (256)} & 27$\times$27(729) & \multicolumn{2}{c}{32$\times$32 (1024)} & \multicolumn{3}{c}{64$\times$64 (4096)} & \multicolumn{2}{c}{81$\times$81 (6561)} \\
\hline
\textbf{\# of Regions:} & 9 & 16 & 9 & 64 & 16 & 256 & 64 & 16 & 81 & 9
\\
\hline
\textbf{Region Size:} & {3$\times$3} & {4$\times$4} & {9$\times$9} & {4$\times$4} & {8$\times$8} & {4$\times$4} & {8$\times$8} & {16$\times$16} & {9$\times$9} & {27$\times$27}
\\
\hline
\textbf{\# of States:} & 9 & 16 & 81 & 16 & 64 & 16 & 64 & 256 & 81 & 729
\\
\hline
\textbf{Q-Table Size:} & 45 & 80 & 405 & 80 & 320 & 80 & 320 & 1280 & 405 & 3645 \\
\hline
\textbf{\# of Iterations:} & 20 & 20 & 20 & 20 & 20 & 20 & 20 & 20 & 20 & 20 \\
\hline
\textbf{\# of Episodes:} & 40 & 40 & 40 & 40 & 40 & 40 & 40 & 40 & 40 & 40\\
\hline
\textbf{\# of Steps:} & 75 & 125 & 600 & 125 & 500 & 125 & 500 & 2000 & 600 & 6000\\ 
\hline
\textbf{Epoch Time(S):} & 0.16955 & 0.27642 & 1.30856 & 0.25871 & 1.08395 & 0.25251 & 1.09411 & 4.28103 & 1.28783 & 13.04665 \\
\hline
\textbf{Iteration Time(S):} & 6.78 & 11.05 & 52.34 & 10.34 & 43.35 & 10.10 & 43.76 & 171.24 & 51.51 & 521.86 \\
\hline
\textbf{Total time:} & 2.26m & 3.68m & 17.4m & 3.44m & 14.4m & 3.36m & 14.58m & 57.08m & 17.17m & 2.89h \\
\bottomrule 
\end{tabular}
}
}
\end{table*}

\subsection{Part-1}
In the first part of the simulation, we evaluated the performance of the proposed method with 5 UAVs in a $81 \times 81$ grid plane with 9 predefined regions. Each region has 27 cells in length and 27 cells in width which means it covers an area of 729 cells that is equal to the number of states for each UAV.

\begin{figure}[b!]
	\centering
.	\includegraphics[width=1\columnwidth]{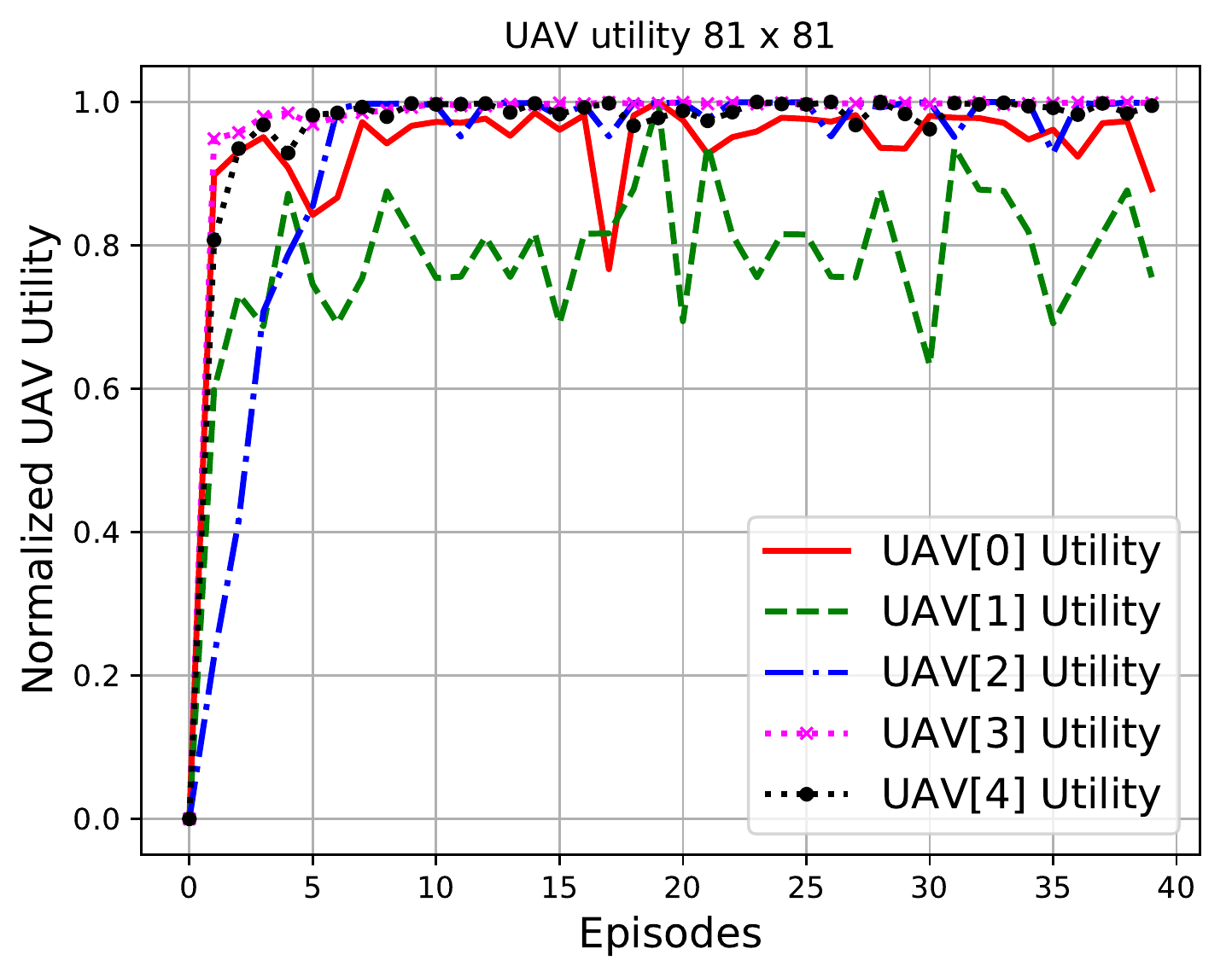}
	\caption{Summation throughput rate for all UAVs.}
    \label{fig:sumutility}
\end{figure}

The sum utility for all UAVs is shown in  Figure \ref{fig:sumutility}. In this sample scenario, UAV[4] is the relay UAV and the rest of them are the sensing UAVs. The summation throughput for each episode is measured which is the accumulated rate for 6000 transmissions during each episode. Based on the plot observation, the UAVs start learning the optimal locations for transmission as the steps proceed. At the final episodes, there are more experienced state-action values in the Q-table for each UAV, as a result, they converge faster to the best or optimal location for maximizing the individual throughput. Since the values of throughput depend also on the location of the allocated regions, the plot shows the normalized rate for UAVs to have better demonstration over the behavior.

\begin{figure}[b!]
	\centering
	\includegraphics[width=1\columnwidth]{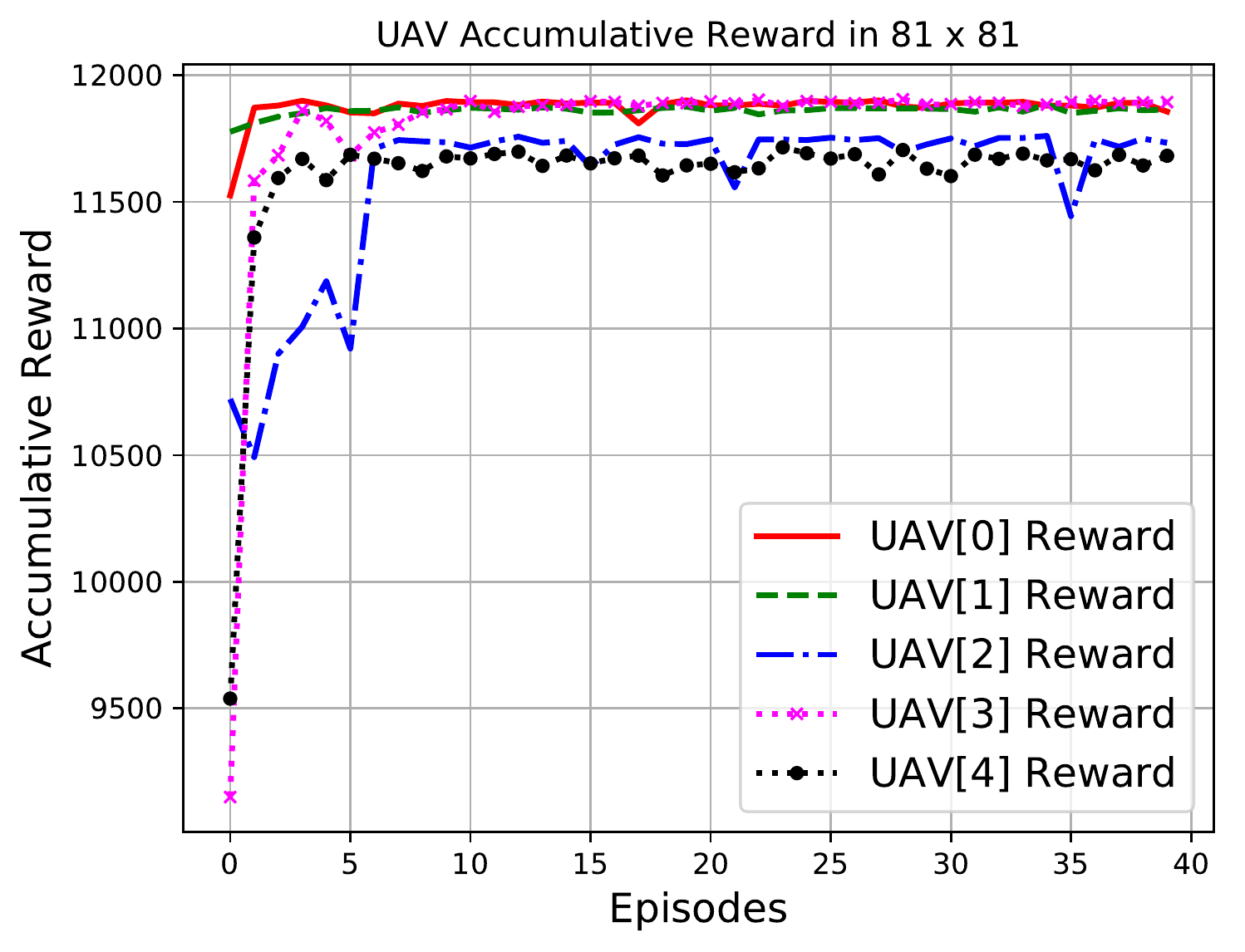}
	\caption{The accumulative reward for all UAV over all episodes}
    \label{fig:reward}
\end{figure}

Figure \ref{fig:reward} demonstrates the accumulative reward during each episode for 5 UAVs. As it is shown, the agents start the learning and the exploring at lower episodes and because of that, they gain more random rewards which can be negative or positive. At the final episodes, they are using the updated Q-values based on the reward and Q-value updating functions in (\ref{eq:reward}) and (\ref{eq:Qupdate}), respectively. Hence, it is more beneficial for the UAVs to merge to the optimal locations based on the energy consumption and the rate differential values between two consecutive steps. Hence, they will earn more positive reward compared to the initial episodes. Also, there is no correlation between the UAVs and they follow the same reward function for updating, as a result, they show the same behavior for the accumulative reward.   

\begin{figure}[bth]
	\centering
	\includegraphics[width=1\columnwidth]{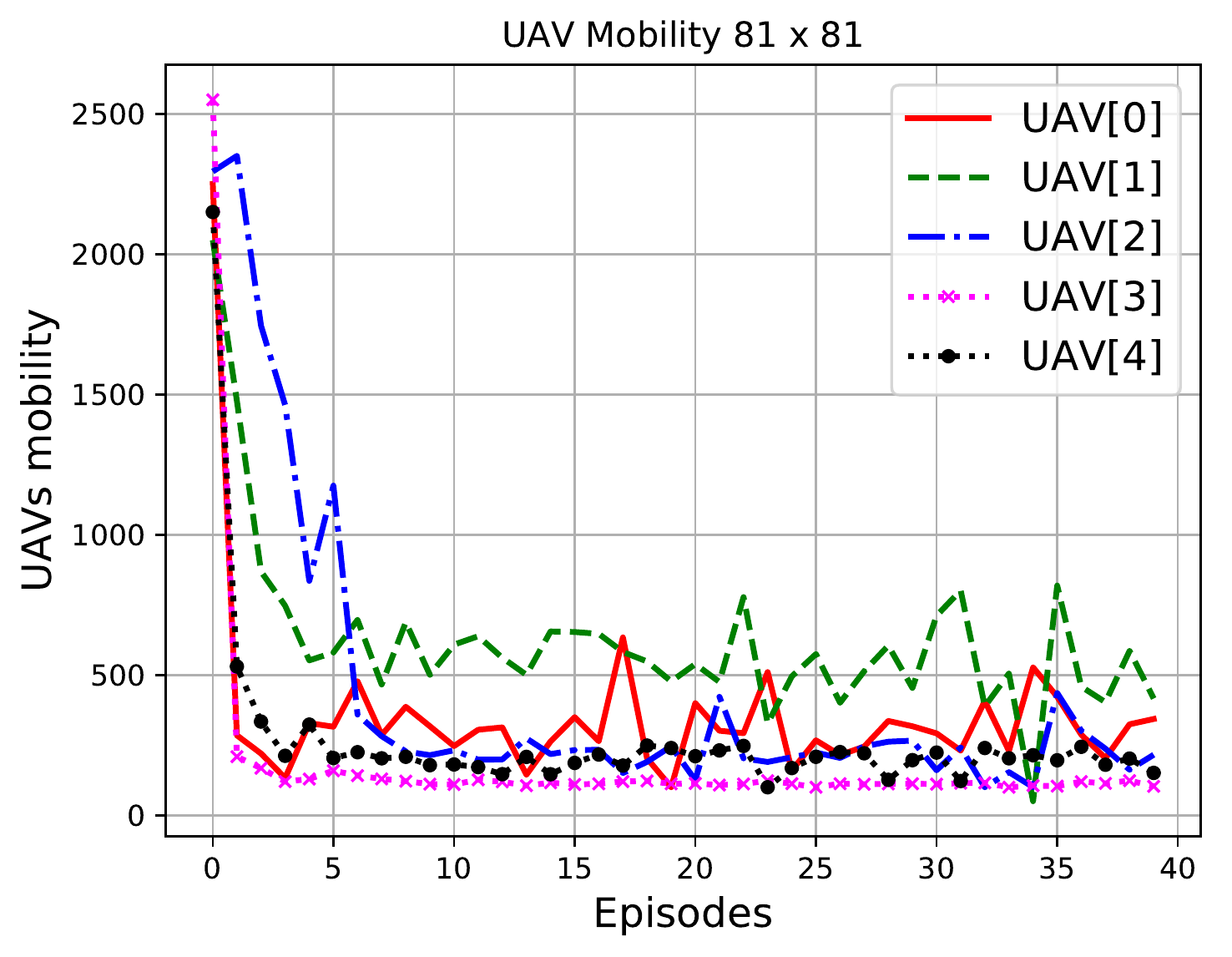}
	\caption{Number of movements for all UAVs in each episode.}
    \label{fig:movement}
\end{figure}

We explicitly defined the energy consideration in the reward function. Hence, the expectation is to have less mobility at the final episodes. Since the mobility costs lots of energy, it brings negative reward for each UAV. The agents learn during action-decisions to have only necessary movements and those necessary moves will result in an optimal location for higher throughput. Based on Figure \ref{fig:movement}, at the beginning of the learning process, the Q-values are initialized with zeros. Hence, there is no real difference between the exploring and the exploiting and the agents have more mobility to get experience with different cells in their region. At the final episodes, when the Q-table is updated with recent values, the UAVs know their best actions to take in order to reach the optimal states. As a result, they are more stable for the final episodes. At the early episodes, most UAVs took around 2000 actions for mobility, while at the final episodes this number is less than 500 which shows their intention to be more stable.

\begin{figure}[hbt]
	\centering
	\includegraphics[width=1\columnwidth]{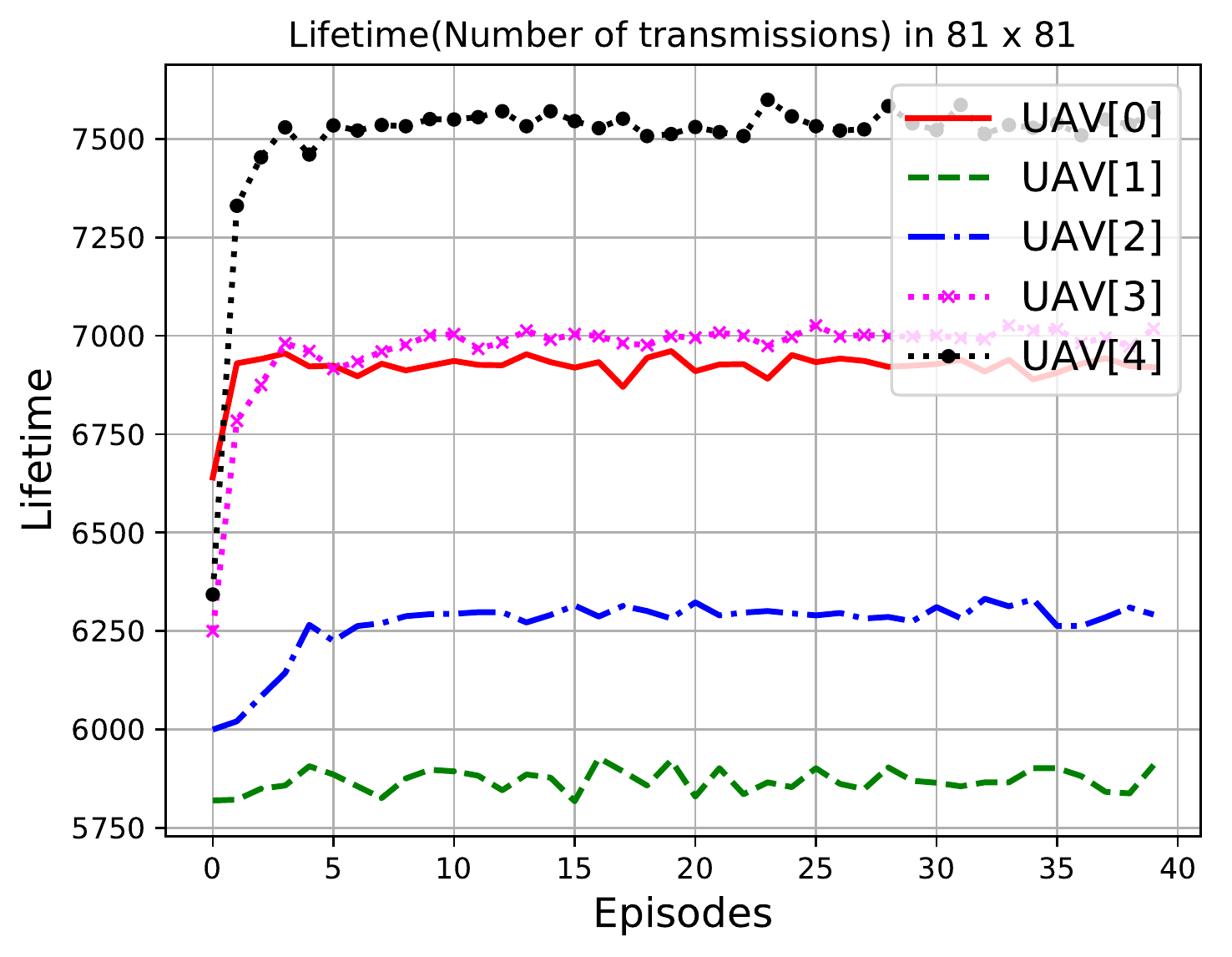}
	\caption{Number of successful transmissions for each UAV over all episodes.}
    \label{fig:lifetime}
\end{figure}

Figure \ref{fig:lifetime} investigates the lifetime for all UAVs based on the number of successful transmissions. If the battery energy is drained completely, then the UAV is considered as a dead node which cannot operate anymore. In this scenario, instead of 6000 steps, the UAVs transmit or relay infinitely to find the maximum number of transmission. Based on Figure \ref{fig:lifetime}, at the early episodes, because of more movements, the UAVs are more prone to higher energy consumption rates and at the final episodes, they are more stable in their locations and the only cost for energy is the transmission. Besides, since the sensing network is borrowing the spectrum from the primary network, the goal is to allocate a UAV which has the longest lifetime to the primary network to utilize the leased spectrum for the critical situation. Based on the emergency center search algorithm, UAV[4] was the most suitable one to relay and it has the best lifetime among all UAVs in the network. 

\begin{figure}[hbt]
	\centering
	\includegraphics[width=1\columnwidth]{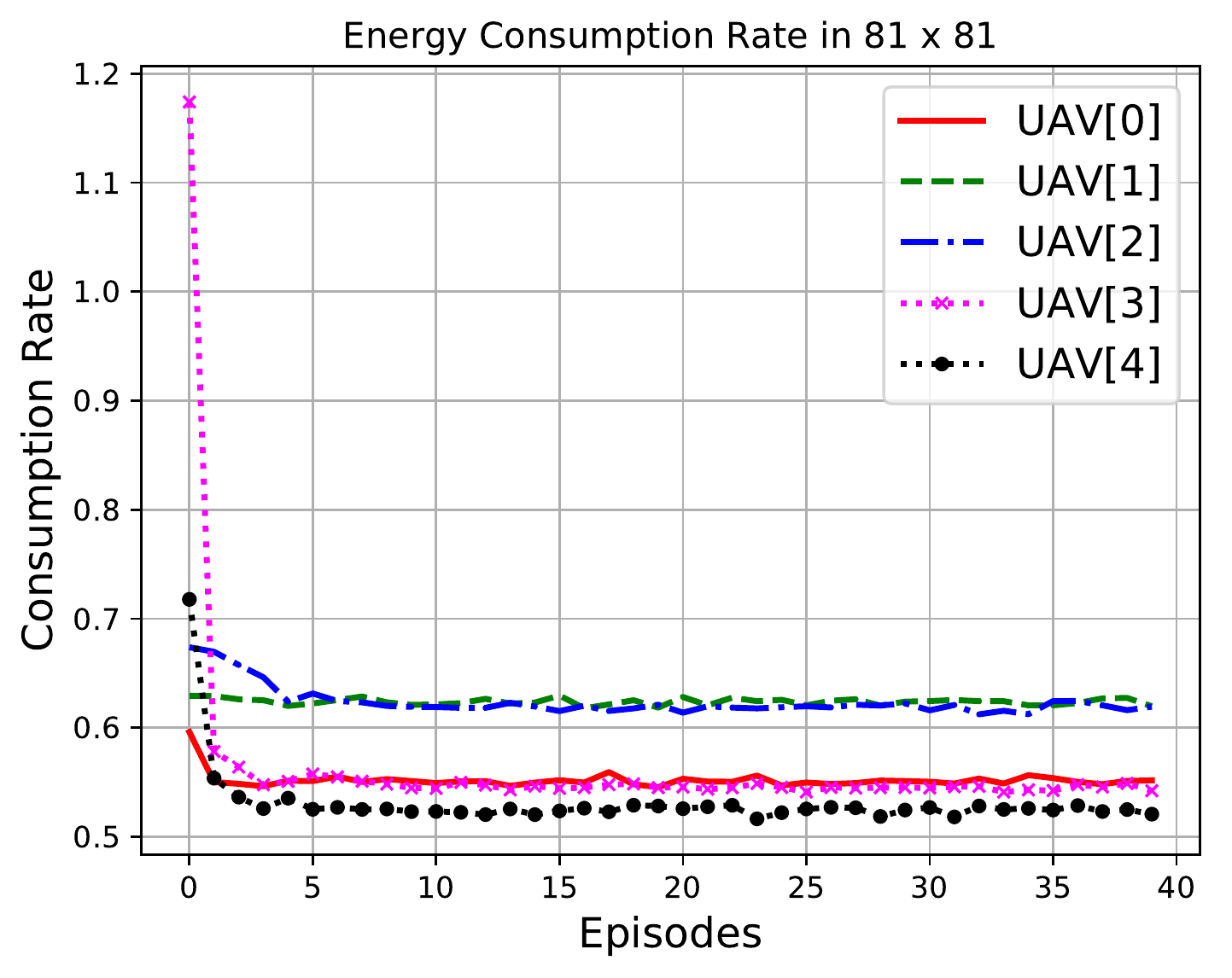}
	\caption{Energy consumption rate for all UAVs in each episode.}
    \label{fig:consumptionrate}
\end{figure}

Figure \ref{fig:consumptionrate} shows the energy consumption rate for all UAVs. To measure the energy consumption rate, we considered the initial energy, the energy at 75\% of all steps in each episode, and the lifetime of the agent in that episode. Based on the figure, UAV[4] has the least energy consumption rate among all UAVs since it performs the task as a relay for the primary (terrestrial) network. The purpose of the searching algorithm for the primary region was to allocate the longest lifetime UAV to the primary network. 

Unlike the exponential behavior in \cite{shamsoshoara2019solution}, the processing time in this work is not increasing at the same pace because of two reasons: first) there is no correlation between the UAVs' strategies and each UAV acts independently, actions and rewards are unique for each UAV, the behavior of one agent does not affect the other ones. Second) increasing the whole grid size does not necessarily increase the processing time since the processing time depends on the number of regions and the regions size. For instance, based on Table. \ref{tab:sim}, the region size of $4 \times 4$ in both grid sizes of $32 \times 32$ and $64 \times 64$ has the same total simulation time to fill up the Q-table. Using the emergency center to allocate the task among the UAVs reduces the converging time significantly.

\subsection{Part-2}
In the second part of the simulation, we compared the performance of the proposed algorithm with four other methods. We considered 5 modes; mode[0] which is the proposed method of this work that the emergency center considers the priority for critical regions and allocates the UAVs to the regions to prolong the lifetime of the network. The UAVs also use the Q-learning algorithm to find the best cell in their own region to have an optimal throughput rate. In mode[1], the emergency center still considers prioritized regions for the UAVs to have optimal throughput. However, the UAV allocation is random, the UAVs still utilize the RL to find the best cell in the regions. In mode[2], both the regions assignment and the UAV allocations are random; but the UAVs use the RL algorithm in each region. Mode[3] is the opposite of mode[0], the regions are chosen randomly not based on the priority, but the emergency center allocates the UAVs wisely based on the energy consumption rate to maximize the network lifetime. The emergency center considers the prioritized regions and allocates the UAVs to the regions in mode[4] based on the two mentioned search algorithms in Section \ref{subsec:Search}. However, the mobility pattern for the UAVs is based on a predefined random path. 

\begin{figure}[hbt]
	\centering
	\includegraphics[width=1\columnwidth]{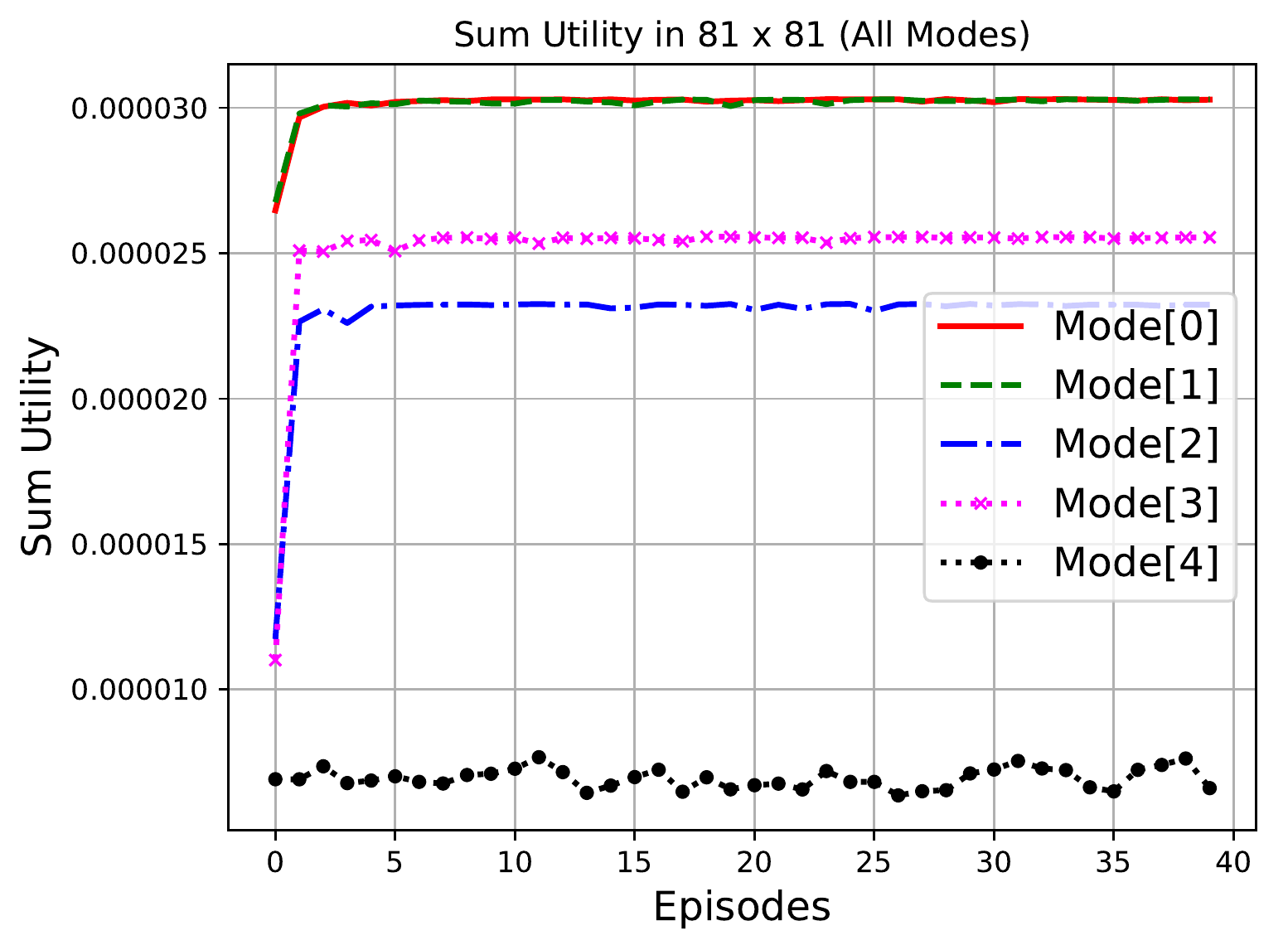}
	\caption{Throughput rate comparison in different conditions, Modes[0-4].}
    \label{fig:Modes_throughput}
\end{figure}

Figure \ref{fig:Modes_throughput} demonstrates the summation utilities for 5 different modes. In cases 0 and 1, the optimal summation throughput is derived when the emergency center considers the critical region as a priority for the UAVs. The emergency center also allocates the best UAVs for the sake of network lifetime in case 0. However, the UAVs are chosen randomly in case 1 which does not affect the throughput. On the other hand, in cases 2 and 3, the throughput is decreased since the regions are chosen randomly still the emergency center allocated the UAVs wisely based on the bipartite graph matching. Obviously, case 4 with all random actions has the worst performance.

\begin{figure}[hbt]
	\centering
	\includegraphics[width=1\columnwidth]{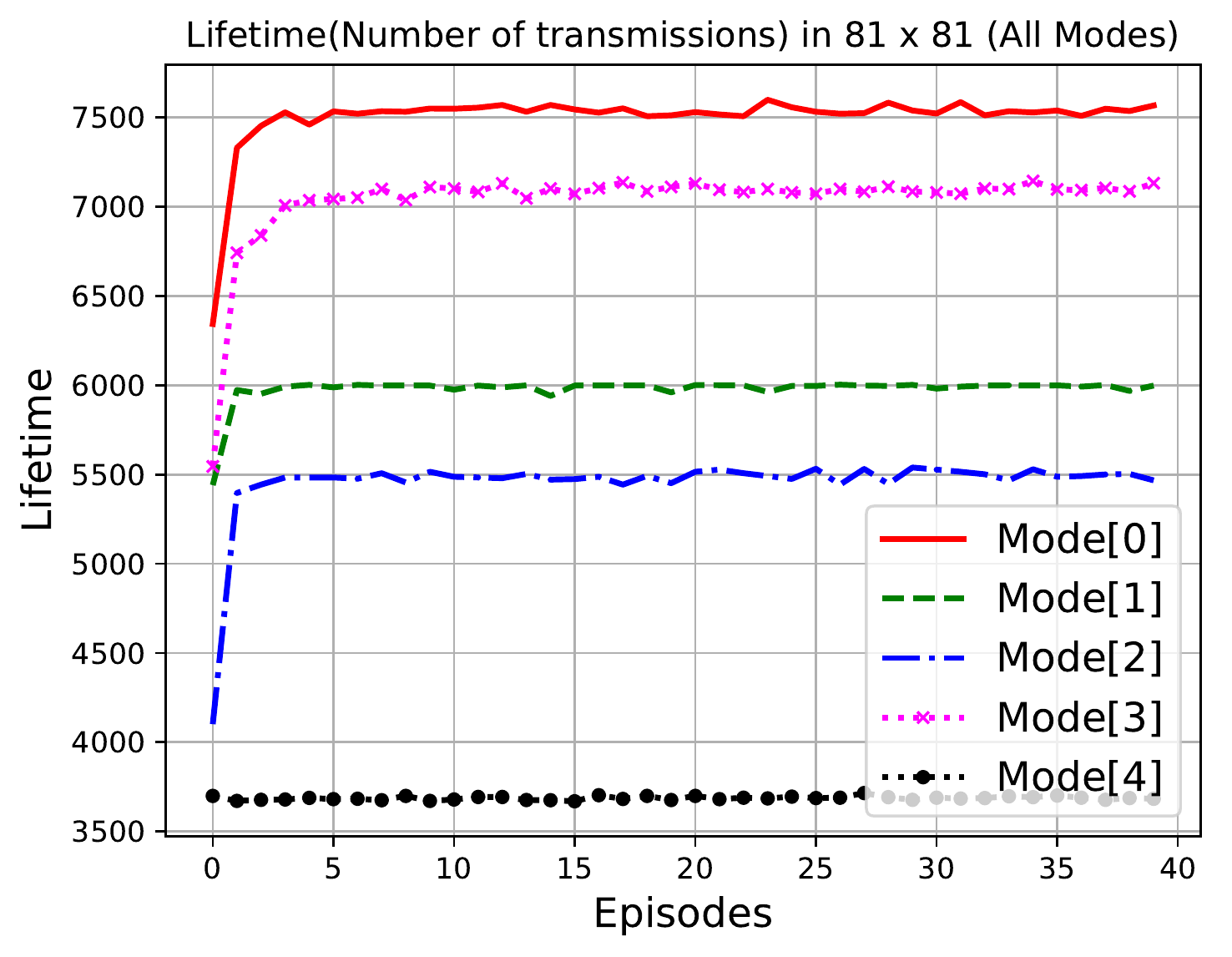}
	\caption{Lifetime comparison in different conditions, Modes[0-4].}
    \label{fig:Mode_lifetime}
\end{figure}

Figure \ref{fig:Mode_lifetime} shows the lifetime for all cases. Case 0, the proposed method, has the longest lifetime in all episodes. To define the lifetime, we consider the relay UAV lifetime, and we assumed that after the battery depletion, the sensing UAVs do not have access to the spectrum anymore. Based on this observation, cases 0 and 3 where the emergency center chooses the optimal UAVs, the lifetime is significantly enhanced compared to other cases. Moreover, choosing the regions also affects the lifetime. It shows that the emergency center's decisions have crucial results on the lifetime rather than the throughput. 

\begin{figure}[hbt]
	\centering
	\includegraphics[width=1\columnwidth]{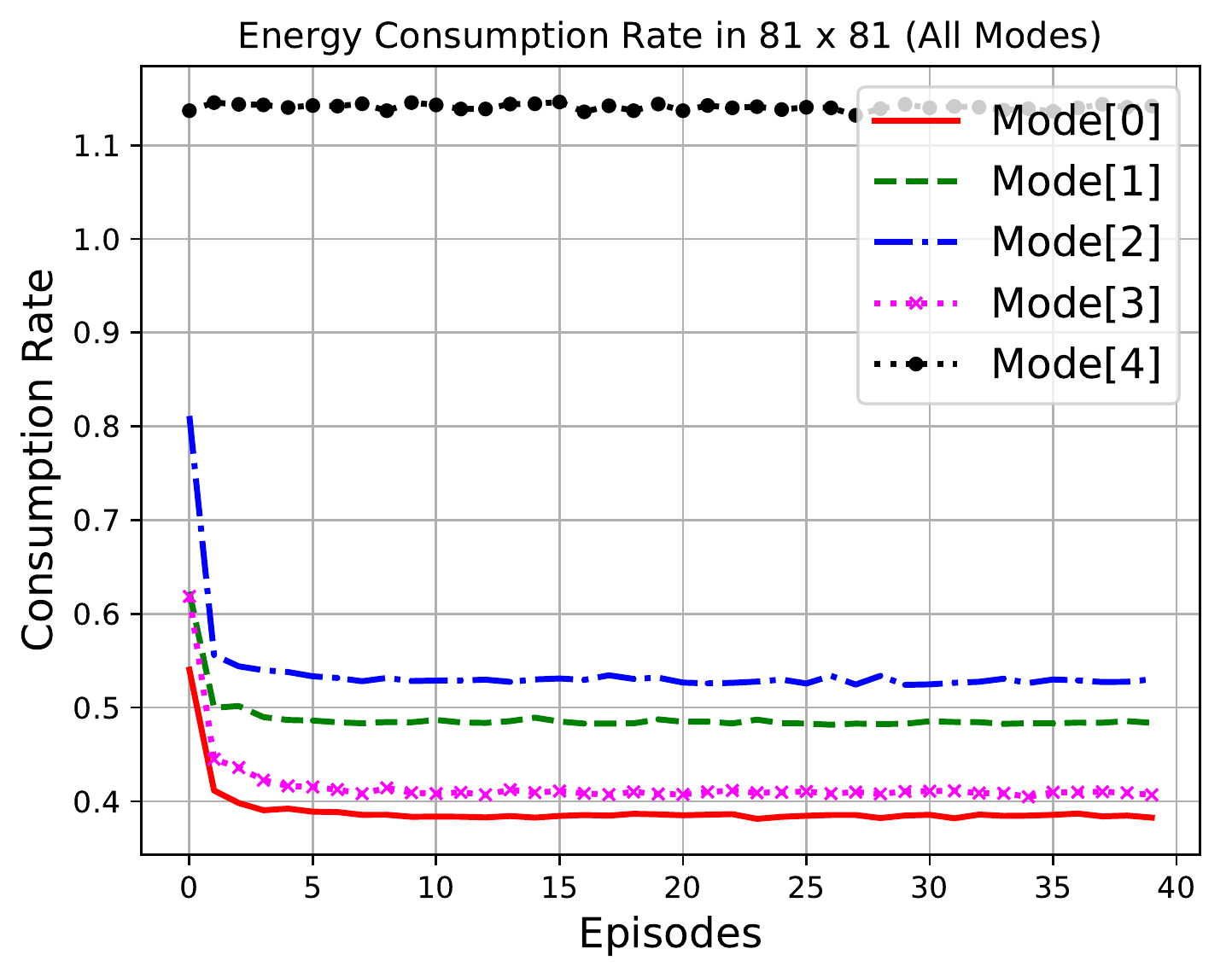}
	\caption{Comparison of energy consumption rate in different conditions, Modes[0-4].}
    \label{fig:Modes_energyconsump}
\end{figure}

For the energy consumption rate, we considered each UAV's initial energy, the energy at the 75\% steps in each episode, and the lifetime of the relay UAV. The average is performed over all the UAVs and the runs. Figure \ref{fig:Modes_energyconsump} shows the proposed method (case 0) has the least significant energy consumption rates for all the UAVs in the network. Case 3 with the random regions and the allocated UAVs follows case 0. Case 1 with the random UAVs and the regions selected by the emergency center is in the third rank and case 2 with the random regions, the random UAV allocation, and the RL algorithm is the fourth one. Case 4 where both the UAVs and the regions are allocated by the emergency center but with a predefined trajectory has the most energy consumption rate.

%% file: Texfiles/conclusion.tex
\section{Conclusion}\label{Conclusion}

In this paper, a disaster relief situation is considered as a sample scenario where a group of UAVs observes critical information such as wildfire situations for an emergency center. The UAV network during such critical missions  may require additional spare spectrum. To address this demand, we developed a spectrum sharing model where one UAV acts as a relay and forwards data for a terrestrial network in exchange for the required spectrum and the rest of the UAVs utilize this spectrum to transmit the sensed information. The emergency center determines the regions for the primary and secondary networks to maximize the throughput and it allocates the UAVs into the chosen regions based on a predefined grid plane to maximize the network lifetime and reduce the energy consumption. A reinforcement learning approach is used for all the UAVs to find the best cell in each region without any prior information of the environment. The simulation results show that after a certain amount of iterations and episodes, the UAVs converge to the optimal state with a fewer number of actions. We also compared the proposed method with other random assignments and allocations. 
For larger gird size planes, it is possible to consider the approximate location of the grid plane for hazardous and dangerous areas and perform offline learning to save the time and come up with a predefined Q-table. Then, the UAVs can act completely greedy based on the Q-table to travel to the best location without any extra action. 